\documentclass[10pt,twocolumn,letterpaper]{article}

\usepackage{wacv}
\usepackage{times}
\usepackage{epsfig}
\usepackage{graphicx}
\usepackage{amsmath}
\usepackage{amssymb}
\usepackage{multirow}
\usepackage{color}
\usepackage[pagebackref=true,breaklinks=true,letterpaper=true,colorlinks,bookmarks=false]{hyperref}

\usepackage{float}

\usepackage[ruled,linesnumbered]{algorithm2e}

\usepackage{amsthm}

\usepackage{subcaption}

\usepackage{xcolor}
\usepackage[export]{adjustbox}

\usepackage{booktabs}

\usepackage{stfloats}

\newcommand\SizeFigAttentionMap{0.11}     
\newcommand\SizeFigCompareHRLarge{0.18}     
\newcommand\SizeFigCompareHRSmall{0.443}    
\newcommand\SizeFigCompareLR{0.138}         

\wacvfinalcopy 

\def\wacvPaperID{****} 

\ifwacvfinal\fi
\begin{document}

\title{Component Attention Guided Face Super-Resolution Network: CAGFace}

\author{Ratheesh Kalarot \\
The University of Auckland \\
{\tt\small rkal018@aucklanduni.ac.nz}
\and
Tao Li \\
Purdue University \\
{\tt\small taoli@purdue.edu}
\and
Fatih Porikli \\
The Australian National University \\
{\tt\small fatih.porikli@anu.edu.au}
}

\maketitle
\ifwacvfinal\thispagestyle{empty}\fi

\begin{abstract}
To make the best use of the underlying structure of faces, the collective information through face datasets and the intermediate estimates during the upsampling process, here we introduce a fully convolutional multi-stage neural network for 4$\times$ super-resolution for face images. We implicitly impose facial component-wise attention maps using a segmentation network to allow our network to focus on face-inherent patterns. Each stage of our network is composed of a stem layer, a residual backbone, and spatial upsampling layers. We recurrently apply stages to reconstruct an intermediate image, and then reuse its space-to-depth converted versions to bootstrap and enhance image quality progressively. Our experiments show that our face super-resolution method achieves quantitatively superior and perceptually pleasing results in comparison to state of the art.
\end{abstract}

\section{Introduction}\label{sec:introduction}

Our brains are wonderfully attuned to perceiving faces. In addition to the visual cortex in the occipital lobe, the entire region of the brain called the fusiform gyrus is dedicated to interpreting and forming a mental representation of faces~\cite{susana2018}. From early childhood, even very shortly after birth, human brains possess facial inference capacities and display more interest in face images than any other pattern~\cite{morton1991}. 
As a species, we almost obsessively monitor and pay close attention to subtle details in paces that can allow gleaning into the origin, emotional state, internal thought process, level of engagement, and health qualities of others around us. Most of us pay more attention to faces than we do to anything other object categories. Supporting this, many gaze tracking studies show that the profile picture or avatar is the first place the eye is drawn to on social media profiles~\cite{todorov2014}. Pictures with human faces are with a large margin more likely to receive likes than the ones with no faces. It is not surprising that almost one-third of social media images are selfies and more than half are tagged with a label relates to face.

\begin{figure}[t]
\begin{center}
    \begin{subfigure}[b]{0.235\textwidth}
        \includegraphics[width=\textwidth]{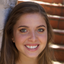}
        \\
        \includegraphics[width=0.455\textwidth,cfbox=red 1pt 1pt]{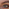}
        \includegraphics[width=0.455\textwidth,cfbox=blue 1pt 1pt]{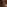}
        \\
        \includegraphics[width=0.455\textwidth,cfbox=red 1pt 1pt]{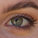}
        \includegraphics[width=0.455\textwidth,cfbox=blue 1pt 1pt]{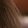}
        \caption{Input LR image.}
    \end{subfigure}
    \begin{subfigure}[b]{0.235\textwidth}
        \includegraphics[width=\textwidth]{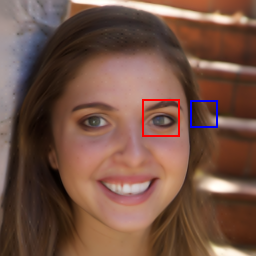}
        \\
        \includegraphics[width=0.455\textwidth,cfbox=red 1pt 1pt]{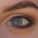}
        \includegraphics[width=0.455\textwidth,cfbox=blue 1pt 1pt]{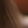}
        \\
        \includegraphics[width=0.455\textwidth,cfbox=red 1pt 1pt]{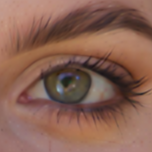}
        \includegraphics[width=0.455\textwidth,cfbox=blue 1pt 1pt]{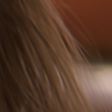}
        \caption{Our SR results.}
    \end{subfigure}
    \vspace{-6mm}
\end{center}
\caption{Our method can 4$\times$ super-resolve face images of any size. 
Top row: 64$\times$64 LR input and our result.
Middle row: enlarged areas from the above images.
Bottom row: enlarged areas when the input LR image is 256$\times$256
(see supplementary for the whole image and its super-resolved counterpart).
Please \textbf{zoom in} for the best view.}
\label{fig:intro}
\end{figure}


The resolution of the faces is an essential factor, and certain features appear to be found more attractive in higher resolution across individuals and cultures~\cite{langois2000,little2011}. Especially the eye and mouth regions are critical for face perception, as well as for neural responses to faces~\cite{smith2009,dehaas2016}. Moreover, the interpretation of facial features is determined by the relative arrangement of parts within a face context~\cite{maurer2002}. Attention selection and guidance, thus, are important elements of high-resolution stimuli in the modeling of the processes in visual processing. 

High-resolution face images provide crucial clues not only for human observation but also for computer analysis~\cite{fasel2003automatic,zhao2003face}. The performance of common facial analysis techniques, such as face alignment~\cite{bulat2017far} and identification~\cite{taigman2014deepface}, degrade when the resolution of a face is low. To provide a viable way to recover a high-resolution (HR) face image from its low-resolution (LR) counterpart, many face super-resolution methods ~\cite{Zhou2015,yu2016ultra,yu2017hallucinating,yu2018eccv,zhu2016deep,cao2017attention,dahl2017pixel} that rely on deep learning networks are proposed in recent years. Some of these methods explore direct image intensity correspondences between LR and HR faces, albeit being limited to low-resolution, e.g., 16$\times$16, input images where the whole face to be included in the image. They can neither handle large input faces due to computational and memory requirements in training and inference times nor can they resolve fine-grained face-specific patterns. Besides, their dependency on near-frontal faces, which is prevalent in popular datasets~\cite{Liu2015faceattributes,LFWTech}, restricts their usage for large pose variations causing distorted facial details. A naive idea to remedy this problem is to augment the training data with large pose variations during the training stage. However, this strategy leads to suboptimal results due to the increased variance of face data to be modeled and also potentially erroneous localization of facial landmarks, which is a difficult task in small LR images under large pose variations.

In this paper, in contrast to previous attempts that often demand and apply the whole face image through their neural layers, we adapt a patch-based face super-resolution method that can operate efficiently on large input faces. Our intuition is that, although it is challenging to detect facial landmarks of the face accurately, it is possible to estimate patch-based attention maps of facial components approximately and steer the super-resolution process with these attention maps to facilitate more natural and accurate resolution enhancement. 


Our model consists of an off-line trained component network and two super-resolution stages. We first segment facial components using a neural network trained off-line. These components can be hair, skin, eyes, mouth, eyebrows, nose, ears, neck, and similar facial regions. In particular, we use three components; hair, skin, and other parts (eyes, mouth, nose, eyebrows, ears) for simplicity. We apply Gaussian smoothing to decrease the sensitivity of component segmentation errors. We multiply the input image pixel-wise with each component heatmaps to obtain heatmap-weighted components, which allows us to impose components as implicit attention priors. We stack the original image and the attention maps into a block. In the training phase, we randomly sample patches from this face-wise block where each patch includes the cropped original image and the corresponding attention maps. The random sampling generates identically sized patches and their augmented (flipped) versions. In testing, we process the LR image patch-wise and aggregate their HR estimations. 

\begin{figure*}[t]
\begin{center}
\includegraphics[width=\textwidth]{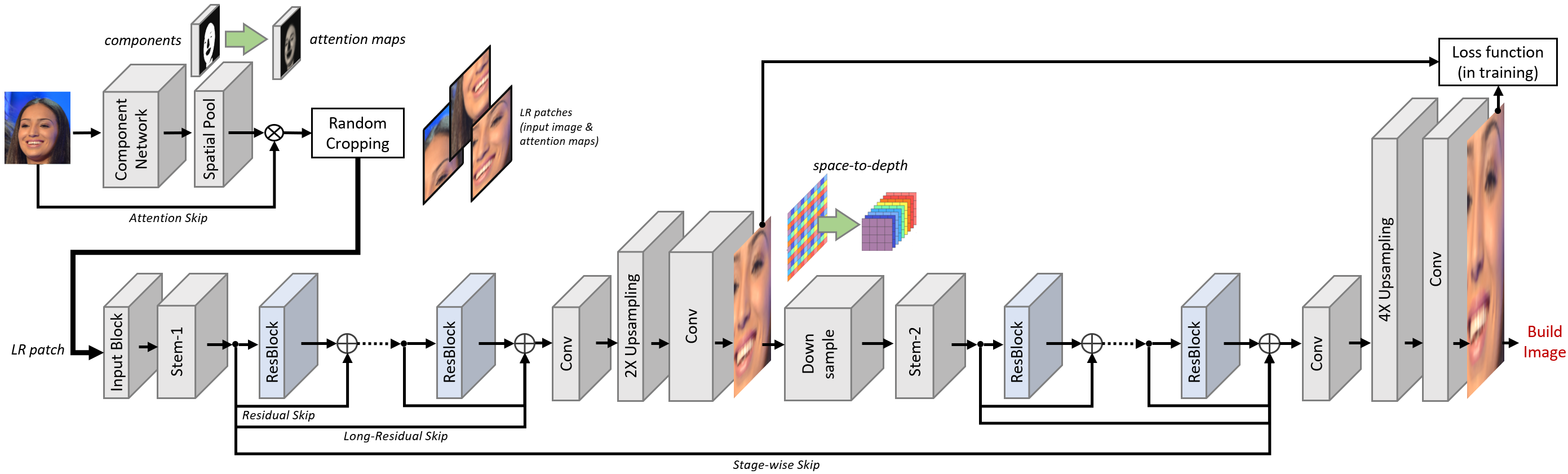}
\end{center}
\caption{CAGFace architecture. First, facial components are segmented, and component-wise attention maps are generated. For training, random patches are sampled. The super-resolution network has two stages; the first stage estimates a 2$\times$ intermediate HR image. The second stage builds on the space-to-depth converted intermediate HR image and uses the original features of the first stem layer through a stage-wise skip-connection while implicitly imposing the component-wise attention. }
\label{fig:architecture}
\end{figure*}

Each super-resolution stage has three main components, as shown in Fig.~\ref{fig:architecture}; a stem layer that blends the input patch channels, a residual backbone that applies fully convolutional blocks on low-resolution feature maps, and a spatial upsampling layer that reconstructs the high-resolution image. The residual backbone is made up of fully convolutional residual units. After a series of residual units, we embed a direct skip connection from the first feature layer to the last one to maintain the influence of the original reference image on the feature map of the last layer. Thus, our backbone is conditioned on reconstructing the residual info, which includes the missing high-resolution patterns in visual data. The residual blocks and direct skip connection also allow us to deepen the backbone, which boosts the overall representation capacity of the network and to increase the areas of the receptive fields for the higher level convolutional layers, which enables better contextual feedback. The residual backbone utilizes the low-resolution image and space-to-depth shuffled estimated high-resolution output of the previous stage, which permits transferring the initial model into progressively more complex networks in the following stages. Note that, each state is an independent network. Following the residual backbone, we apply spatial upsampling layers to reconstruct a higher-resolution image from its feature map. These layers use pixel shuffling with learned weights; therefore, we do not require deconvolutions. The residual backbone prepares the best possible feature maps, which have a large number of channels, and the spatial upsampling layers rearrange these feature maps into the high-resolution images using the learned weights of the filters of these layers.  


To summarize, the contributions of this paper are: 
\vspace{-1mm}
\begin{itemize}
\item  We introduce a patch-based, fully convolutional network for single image face super-resolution that processes patches in their original low-resolution throughout its backbone and layers then reconstructs the high-resolution output from rearranged feature maps.        
\vspace{-2mm}

\item We recurrently apply the super-resolution stages to leverage on the reconstructed high-resolution outputs from the previous stage to enhance estimated high-resolution details progressively. 
\vspace{-2mm}

\item As our experiments demonstrate, our method outperforms existing face super-resolution methods by a large margin without inducing perceptual artifacts. 

\end{itemize}

\section{Related Work}\label{sec:literature}

Image super-resolution aims at restoring an HR counterpart of a given an LR image as input. This task has been one of the most fundamental challenges in computer vision, and many approaches have been proposed within the last two decades including kernel interpolations~\cite{lehmann1999survey}, edge statistics~\cite{fattal2007image, sun2008image}, patch-based  schemes~\cite{freeman2002example, wang2005patch, irani2009super, freedman2011image, yang2012coupled, yang2013fast, huang2015single, sajjadi2017enhancenet}, Bayesian methods~\cite{sun2003image,kim2010single,shan2008fast}, and supervised learning~\cite{aly2005image, ni2007image, zhang2010non}. An in-depth discussion of the available solutions can be found in recent surveys~\cite{yue2016,hayat2018,yang2019}. 

With the compelling advance of deep learning models, in particular, the generative adversarial networks (GAN)~\cite{goodfellow2014generative}, a new wave of convolutional neural network (CNN) based image super-resolution methods have also been proposed. Most notably, SRCNN~\cite{dong2015image} and SRGAN~\cite{ledig2017photo} apply a CNN and a GAN, respectively, to hallucinate HR image pixels. The work in \cite{lai2017deep} progressively estimates the residual of high-frequency details using a Laplacian pyramid super-resolution network, \cite{wang2018recovering} introduces the SFT-GAN for class-conditioned image super-resolution, and \cite{wang2018esrgan} proposes ESRGAN that leveraged a relativistic GAN~\cite{jolicoeur2018relativistic} to estimate the distance between two images. Unlike most super-resolution models that are trained using synthetic LR images, \cite{zhang2019zoom} obtains LR-HR image pairs by zooming-in and -out camera lens to characterize the imaging system degradation functions. We refer readers to \cite{yang2019deep} for a comprehensive overview of deep learning-based and to \cite{yang2014single} for canonical super-resolution approaches.

Many super-resolution methods employ facial priors to achieve higher-resolution faces. Earlier methods assume faces are in a controlled environment, and the variations are minor. For instance, the work proposed in \cite{baker2000hallucinating} uses a spatial gradient distribution as a prior for the frontal faces. In \cite{wang2005hallucinating}, a mapping between LR and HR faces is modeled by an eigen-transform. \cite{kolouri2015transport} learns a nonlinear Lagrangian model for HR face images by finding the model parameters that best fit the LR image. The work in \cite{yang2013structured} incorporates face priors by mapping specific facial components (similar to our method), yet the correspondence between the components is explicitly based on landmark detection, which is difficult to obtain when the upsampling factor is large. The cascaded framework proposed by~\cite{zhu2016deep} super-resolves tiny faces by alternatively optimizing for face hallucination and dense correspondence field estimation. The method presented in \cite{song2017learning} generates facial parts by CNNs and explicitly synthesizes fine-grained facial structures through part enhancement. FSRNet~\cite{FSRNet2018} computes facial landmark heatmaps and alignment parsing maps for end-to-end training.

The imposed image quality measure and the terms of the loss function between the reconstructed and original HR images play a critical role in super-resolution. Peak Signal-to-Noise Ratio (PSNR) is the most common metric to measure the quality~\cite{hore2010image}; however, a higher PSNR value does not necessarily imply a more visually appealing result~\cite{wang2009mean}. To better simulate the human visual perception, Structural Similarity Index Measure (SSIM)~\cite{wang2002universal} separates the task of similarity measurement into three components: luminance, contrast, and structure. Multi-Scale Structural Similarity Index Measure (MS-SSIM)~\cite{wang2003multiscale} adapts to the variations of viewing conditions, and Feature Similarity Index (FSIM)~\cite{zhang2011fsim} extends SSIM to feature space. Inception Score (IS)~\cite{salimans2016improved} measures the quality of generated images and their diversity, and Fr\'echet Inception Distance (FID)~\cite{heusel2017gans} extracts features from an intermediate layer of an Inception Network~\cite{szegedy2016rethinking}. 

Accordingly, many loss functions have been proposed to train deep neural networks for super-resolution, such as pixel-wise mean squared error (MSE)~\cite{van2006image}. While the MSE results in higher PSNR values, it often causes blur and suppresses sharp textures~\cite{wang2009mean}. To overcome this,  perceptual loss~\cite{johnson2016perceptual} imposes feature similarity between the super-resolved and LR images. Perceptual loss is computed over the layer right before the FC layers of VGG19 in \cite{ledig2017photo}, or the B1, B2, and B3 blocks of ResNet50 in \cite{bulat2018super}. A heatmap loss is proposed to preserve structural consistency between LR and HR images further~\cite{bulat2018super}. Leveraging an adversarial loss from a discriminator has been shown to generate convincing results~\cite{ledig2017photo,wang2018esrgan,nah2019ntire} as well.

\section{Proposed Method: CAGFace}\label{sec:method}

Our face super-resolution solution is composed of multiple stages. Here, we use two consecutive stages that achieve 4$\times$ super-resolution, yet our methodology can be applied recurrently for higher upsampling goals. As aforementioned, we bootstrap the super-resolution process by space-to-depth rearranging the estimated high-resolution image into multiple low-resolution channels, imposing the feature maps of the first stem layer (explained below) via a stage-wise skip connections for additional regularization, and applying a second stage network. Our patch-recurrent approach progressively bootstraps on the estimated high-resolution results, thus provides additional performance improvements. 

First, we use a network that segments the facial components. We apply a layer that imposes spatial attention by multiplying the LR input image by the component heatmaps. After a random sampling of patches, we apply two stages of super-resolution networks. 

We achieve 4$\times$ super-resolution with two consecutive stages of 2$\times$ resolution enhancing networks. Notice that, unlike existing methods, our method does not employ a 2$\times$ super-resolution, followed by a second 2$\times$ super-resolution on the output of the first stage. The spatial size of the input feature map to the second stage is \textbf{identical} to the size of the original LR image. We learn the most useful features for 4$\times$ super-resolution after the first stage that also reconstructs a 2$\times$ image. This mid-stage reconstruction enables us to provide an additional regularization for our loss function. 

Each stage contains a separate stem layer, a collection of multiple residual blocks, and an upsampling layer followed by a final mixing layer, as illustrated in Figure~\ref{fig:architecture}. Please see Table~\ref{table:config_table} for the network parameters. In addition to these, the second stage has a depth-to-image conversion layer. These stages have similar kernels, yet their hyperparameters are different. Our network has conventional residual block skip connections and also a stage-wise skip connection that propagates the original features after the first stem layer to just before the final 4$\times$ upsampling layers, as well as a skip-connection over the component network that imposes attention priors on the input image. In testing, we process the LR image patch-wise and aggregate their HR estimations. 

\begin{figure}[t]
\begin{center}
    \begin{subfigure}[b]{\SizeFigAttentionMap\textwidth}
        \includegraphics[width=\textwidth]{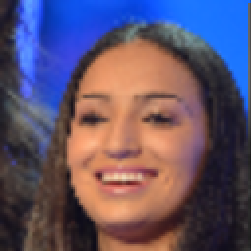}
        \includegraphics[width=\textwidth]{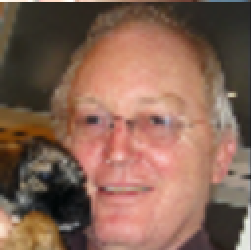}
        \caption{Input}
    \end{subfigure}
    \begin{subfigure}[b]{\SizeFigAttentionMap\textwidth}
        \includegraphics[width=\textwidth]{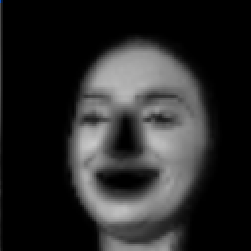}
        \includegraphics[width=\textwidth]{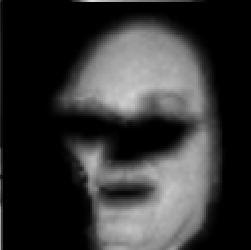}
        \caption{Skin}
    \end{subfigure}
    \begin{subfigure}[b]{\SizeFigAttentionMap\textwidth}
        \includegraphics[width=\textwidth]{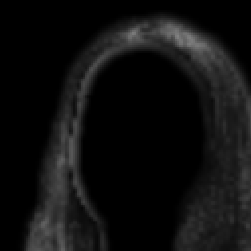}
        \includegraphics[width=\textwidth]{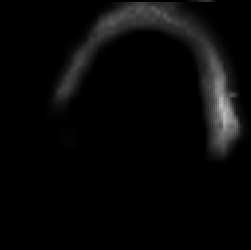}
        \caption{Hair}
    \end{subfigure}
    \begin{subfigure}[b]{\SizeFigAttentionMap\textwidth}
        \includegraphics[width=\textwidth]{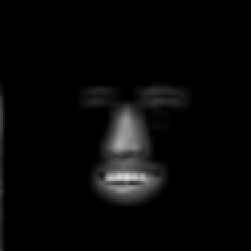}
        \includegraphics[width=\textwidth]{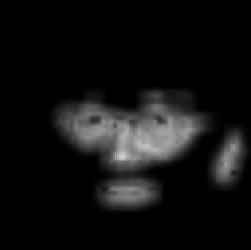}
        \caption{Other parts}
    \end{subfigure}
\end{center}
\vspace{-5mm}
\caption{Sample attention maps from component network.}
\label{fig:attemtionmaps}
\end{figure}

\subsection{Component Network}

For segmentation of the facial components, we followed a similar model that to BiSeNet~\cite{Bisenet2018} that is developed initially for generic purpose pixel labeling such as segmentation of Cityscapes images. BiSeNet has spatial and context paths that are devised to handle the loss of spatial information and shrinkage of the receptive fields, respectively. The spatial path has three convolution layers to obtain a smaller feature map. Context path appends a global average pooling layer on the tail of the Xception network~\cite{chollet2017}. We fine-tuned this model on the CelebAMask-HQ dataset, which has 30,000 high-resolution (1024$\times$1024) face images selected from the CelebA-HQ. Each image has a 512$\times$512, manually-annotated, binary segmentation mask and 19 facial attributes such as skin, nose, eyes, eyebrows, ears, mouth, lip, hair, hat, eyeglass, earring, necklace, neck, and cloth. 

We apply a spatial pooling layer that employs fixed Gaussian spatial kernels to suppress the segmentation errors by providing smoothing. This layer also allows higher values in attention maps to be assigned to more confidently segmented pixels. Finally, we multiply the input image with three spatially pooled components to obtain three gray-level attention maps. Sample attention maps are shown in Figure~\ref{fig:attemtionmaps}. We stack the original LR image and the attention maps into a block, which is our approach to administer attention to the input image. We steer the remaining of our super-resolution network using these maps as attention priors. In the training phase, we randomly sample patches from this block. Each patch, as a result, has the cropped original image and the corresponding attention maps. The random sampling generates identically sized patches and their augmented (flipped in 6 ways) versions. 

\subsection{Stem Layer}

The layer takes a patch block as the input tensor and applies convolutional filters on it. Each depth-wise channel is a color channel of the LR image and corresponding heatmap-weighted components, which are normalized to [-1,1] for efficient backpropagation. The stem layer in the first stage arranges the patch block in a 6-channel tensor. It then applies 256 filters, 3$\times$3$\times$6 each. For the following stage, we have additional channels. After the first stage, we estimated a 2$\times$ super-resolved HR image; thus, we first rearrange (space-to-depth) the pixels of an estimated HR image into 4 LR images. We then combine these LR images into a 12-channel tensor. Notice that we do not impose the heatmaps explicitly again.

Our network uses the original LR resolution frames in all its layers and stages. Since we use the same image size for all layers (except the upsampling layers), the learning becomes more efficient. Multiple references provide spatially vibrant local patterns.

\subsection{Residual Backbone}

The residual backbone applies fully convolutional blocks on low-resolution feature maps generated by the stem layers. It is made up of 16 fully convolutional residual units. Each residual unit has a front convolutional layer followed by a ReLU and a second convolutional layer with a skip connection from the first one. Similarly, the residual backbone has also a direct skip connection from the input to the last residual block. This skip connection allows our network to learn the missing high-resolution details by reconstructing the residual info. The structure of the residual backbone of each stage is identical. The residual blocks and the direct skip connection also permits deepening the residual backbone for each stage. This boosts the overall capacity and increases the receptive field sizes. Thus, residual backbone feature maps have better access to contextual information.  

\subsection{Spatial Upsampling}

We apply spatial upsampling layers to reconstruct a higher-resolution image from the feature map of the residual backbone. Since we shuffle pixels and we apply a set of convolutional filters, our upsampling does not require deconvolution operations. We rearrange the comparably large number of feature map channels per pixel into a high-resolution image using the learned weights of the filters of the upsampling layers. We set the number of layers for the first stage and the second stage to 4 and 5 as the second stage feature map has to generate pixels. Each stage provides 2$\times$ super-resolution, yet it is possible to set the upsampling factor to larger ratios since there the feature maps are sufficiently deep.  

For the goal of higher PSNR results, MSE would be the ideal loss function. However, MSE heavily penalizes the outliers. Recently, the work in \cite{lim2017enhanced} empirically demonstrated that the mean absolute error (MAE) works better than the MSE. In our experiments, we also made a similar observation. In particular, at the initial stages of the training, using the MSE based loss functions caused instability. However, MAE-based loss at the later epochs converges slowly. Therefore, we opted to impose the Huber loss function, which is differentiable and combines the benefits of the MAE and MSE. It is defined as
\begin{equation}
L_{\delta } ( d ) = \left\{ \begin{array}{rl}
 {\frac  {1}{2}}{ d^{2}  } &\mbox{ for $ |  d | \leq \delta ,$} \\
  \delta | d |  - \frac  {\delta^2}{2}  &\mbox{ otherwise}
       \end{array} \right.
\end{equation}
where 
\begin{equation}
 d =  I_{HR}(x,y)  - \hat{I}_{HR}(x,y)  
\end{equation}
is the pixel-wise difference between the target (ground-truth) HR image $I_{HR}$ and the estimated HR image $\hat{I}_{HR}$. Above, we set {$\delta =1$}, which is the point where the Huber loss function changes from quadratic to linear.

\begin{table}[t]
\centering
\vspace{1mm}
\small
\begin{tabular}{|p{25mm}|r|r|}
\hline
Subnetwork & Kernel shape &  Kernel params (bias) \\
\hline 

Stem 1 & 3x3x6x256 & 13824 (256) \\
\hline

\multirow{2}{25mm}{Backbone 1: 16$\times$ \\ ResBlocks (2 layer)} & 3x3x256x256 & 9437184 (4096) \\
                                             & 3x3x256x256 & 9437184 (4096) \\
\hline

\multirow{4}{25mm}{Spatial Upsampling 1} & 3x3x256x256 & 589824 (256) \\
                                 & 3x3x256x1024 & 2359296 (1024) \\
                                 & 3x3x256x3 & 6912 (3) \\
                                 & 3x3x12x256 & 27648 (256) \\
\hline 

Stem 2 & 3x3x256x256 & 587520 (256) \\
\hline

\multirow{2}{25mm}{Backbone 2: 16$\times$ \\ ResBlocks (2 layer)} & 3x3x256x256 & 9437184 (4096) \\
                                             & 3x3x256x256 & 9437184 (4096) \\
\hline 

\multirow{5}{25mm}{Spatial Upsampling 2} & 3x3x256x256 & 589824 (256) \\
                                 & 3x3x512x2048 & 9437184 (2048) \\
                                 & 3x3x512x2048 & 9437184 (2048) \\
                                 & 3x3x512x3 & 13824 (3) \\
                                 & 3x3x3x3 & 81 (3) \\
\hline 
\multicolumn{2}{|l|}{Total trainable parameters in Stage 1} & 21881859\\ 
\hline 
\multicolumn{2}{|l|}{Total trainable parameters in Stage 2} & 38952791 \\ 
\hline
\end{tabular}
\caption{CAGFace network parameters.}
\label{table:config_table}
\end{table}

We trained the first stage and then the second stage by using the learned first stage parameters for initialization. 

\begin{figure*}[t]
\begin{center}
    \begin{subfigure}[b]{\SizeFigCompareHRLarge\textwidth}
        \includegraphics[width=\textwidth]{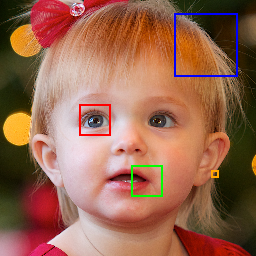}
        \\
        \includegraphics[width=\SizeFigCompareHRSmall\textwidth,cfbox=red 1pt 1pt]{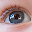}
        \includegraphics[width=\SizeFigCompareHRSmall\textwidth,cfbox=blue 1pt 1pt]{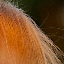}
        \\
        \includegraphics[width=\SizeFigCompareHRSmall\textwidth,cfbox=green 1pt 1pt]{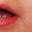}
        \includegraphics[width=\SizeFigCompareHRSmall\textwidth,cfbox=orange 1pt 1pt]{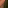}
        \caption{Input \\ (PSNR / SSIM)}
        \label{fig:input}
    \end{subfigure}
    ~
    \begin{subfigure}[b]{\SizeFigCompareHRLarge\textwidth}
        \includegraphics[width=\textwidth]{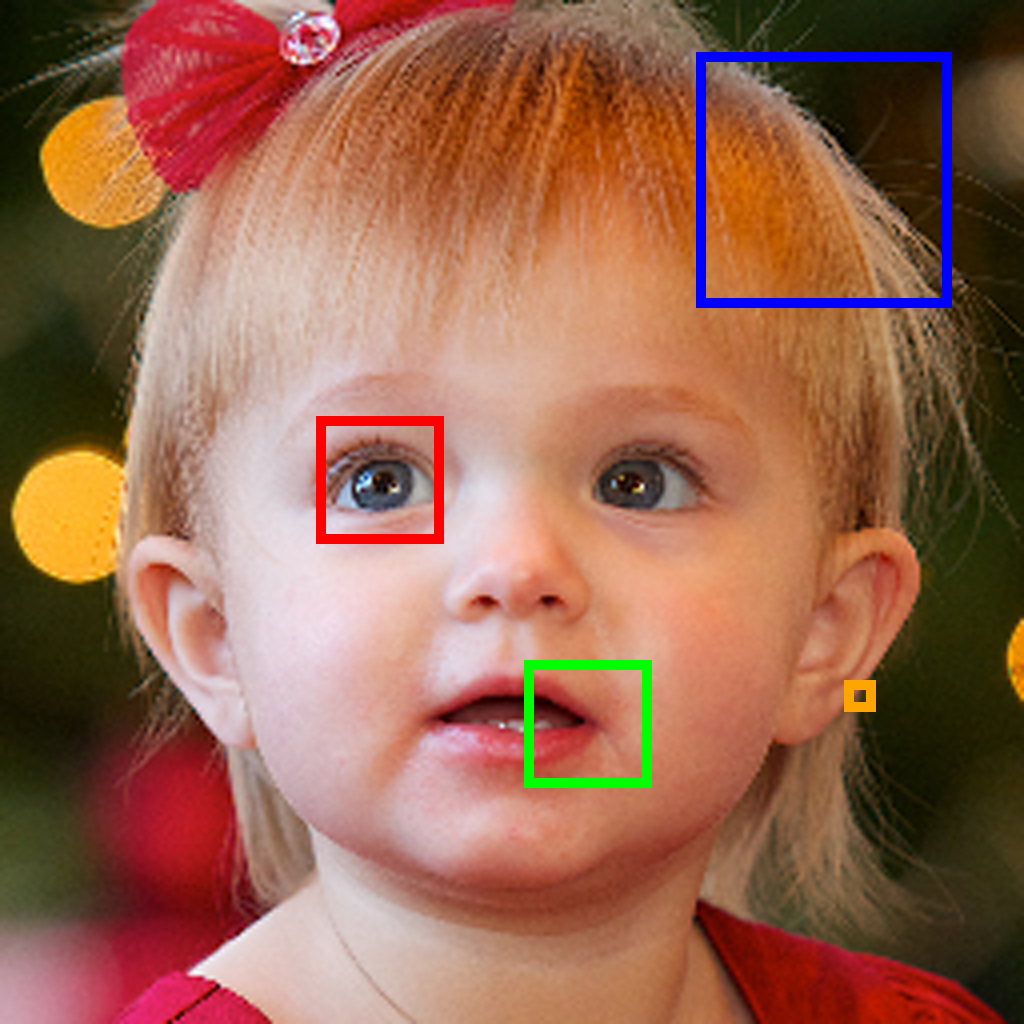}
        \\
        \includegraphics[width=\SizeFigCompareHRSmall\textwidth,cfbox=red 1pt 1pt]{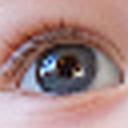}
        \includegraphics[width=\SizeFigCompareHRSmall\textwidth,cfbox=blue 1pt 1pt]{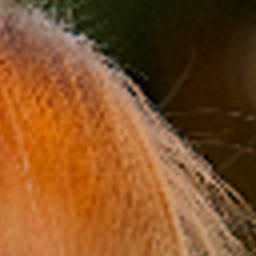}
        \\
        \includegraphics[width=\SizeFigCompareHRSmall\textwidth,cfbox=green 1pt 1pt]{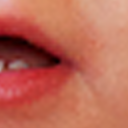}
        \includegraphics[width=\SizeFigCompareHRSmall\textwidth,cfbox=orange 1pt 1pt]{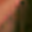}
        \caption{Bicubic \\ (30.96 / 0.830)}
        \label{fig:bicubic}
    \end{subfigure}
    ~
    \begin{subfigure}[b]{\SizeFigCompareHRLarge\textwidth}
        \includegraphics[width=\textwidth]{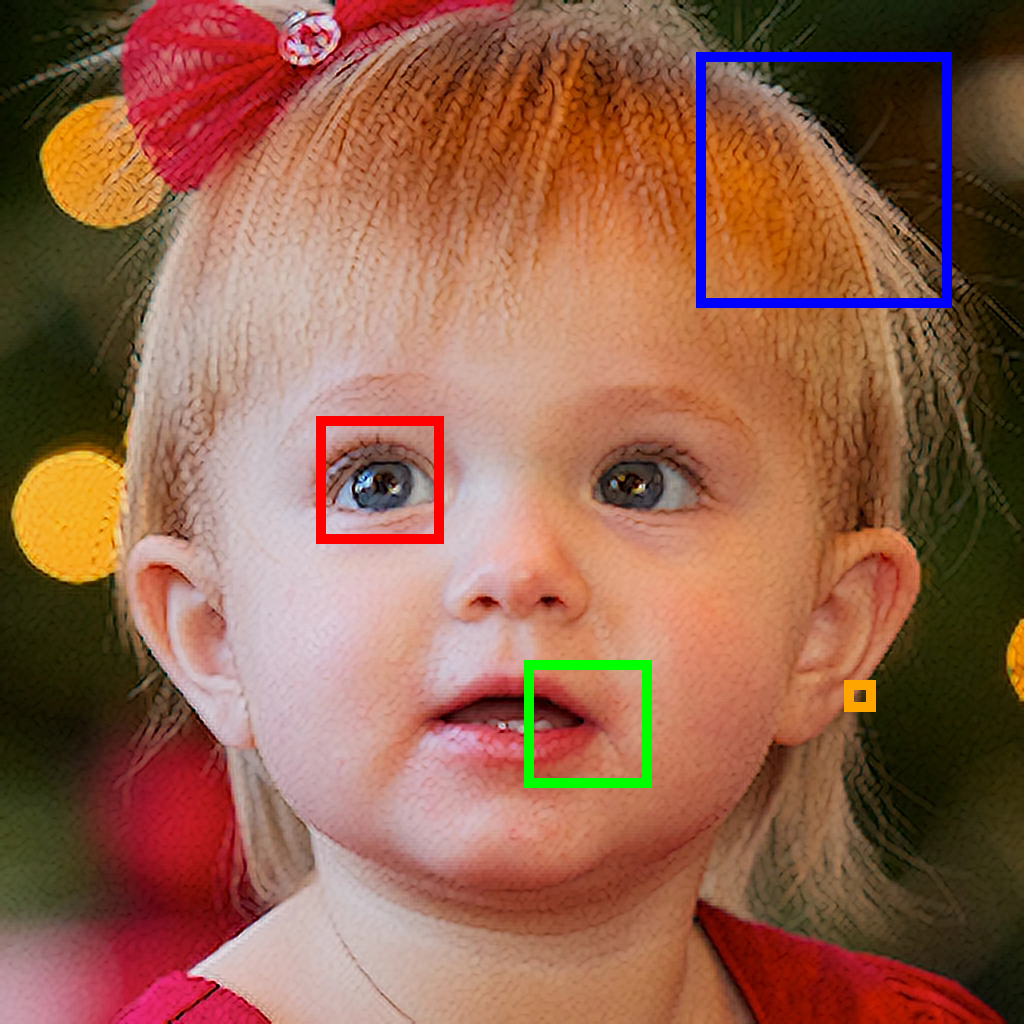}
        \\
        \includegraphics[width=\SizeFigCompareHRSmall\textwidth,cfbox=red 1pt 1pt]{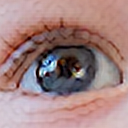}
        \includegraphics[width=\SizeFigCompareHRSmall\textwidth,cfbox=blue 1pt 1pt]{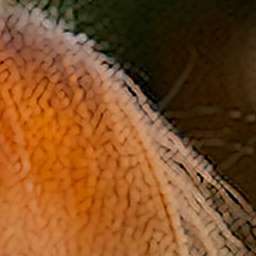}
        \\
        \includegraphics[width=\SizeFigCompareHRSmall\textwidth,cfbox=green 1pt 1pt]{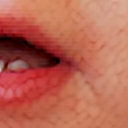}
        \includegraphics[width=\SizeFigCompareHRSmall\textwidth,cfbox=orange 1pt 1pt]{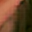}
        \caption{SRCNN~\cite{dong2015image} \\ (27.54 / 0.750)}
        \label{fig:SRCNN}
    \end{subfigure}
    ~
    \begin{subfigure}[b]{\SizeFigCompareHRLarge\textwidth}
        \includegraphics[width=\textwidth]{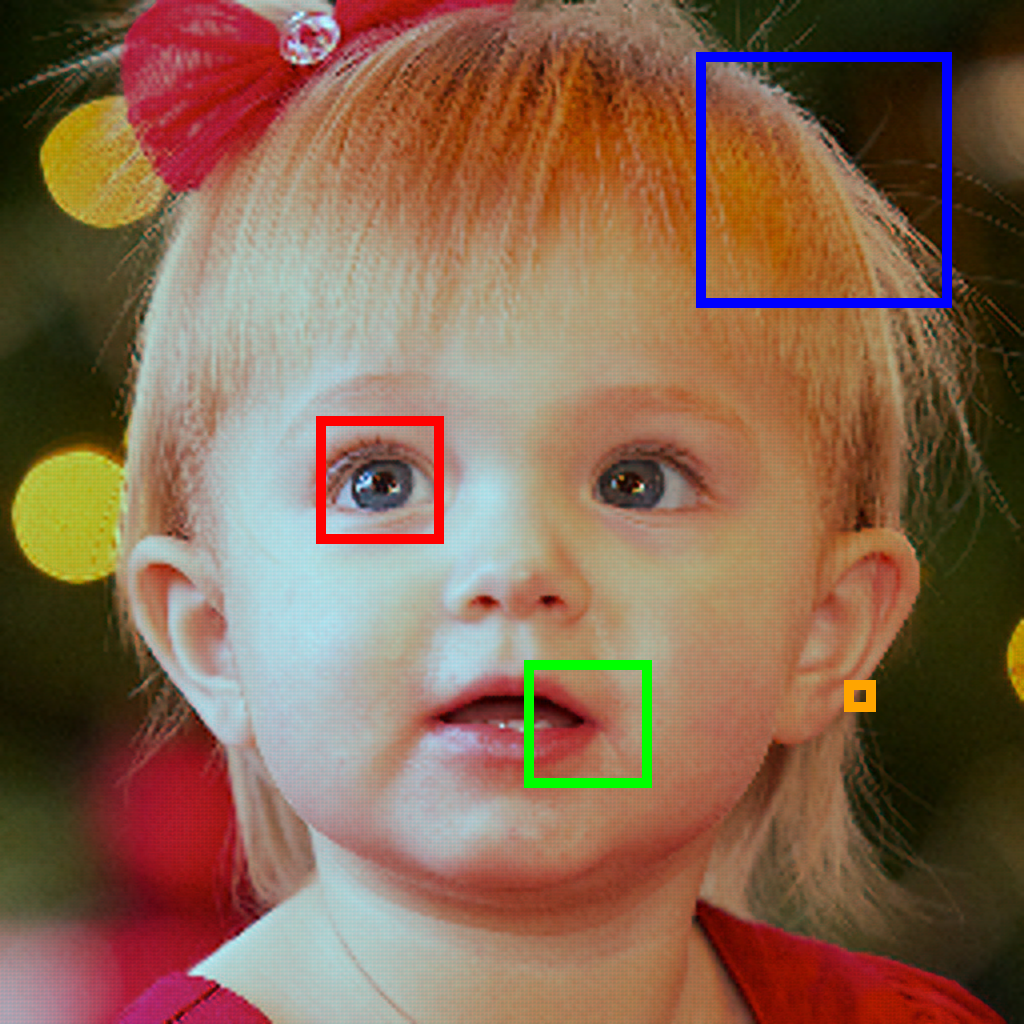}
        \\
        \includegraphics[width=\SizeFigCompareHRSmall\textwidth,cfbox=red 1pt 1pt]{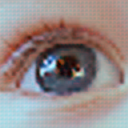}
        \includegraphics[width=\SizeFigCompareHRSmall\textwidth,cfbox=blue 1pt 1pt]{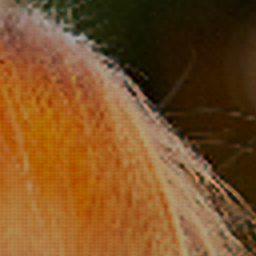}
        \\
        \includegraphics[width=\SizeFigCompareHRSmall\textwidth,cfbox=green 1pt 1pt]{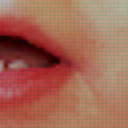}
        \includegraphics[width=\SizeFigCompareHRSmall\textwidth,cfbox=orange 1pt 1pt]{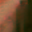}
        \caption{FSRCNN~\cite{dong2015image} \\ (23.56 / 0.749)}
        \label{fig:FSRCNN}
    \end{subfigure}
    ~
    \begin{subfigure}[b]{\SizeFigCompareHRLarge\textwidth}
        \includegraphics[width=\textwidth]{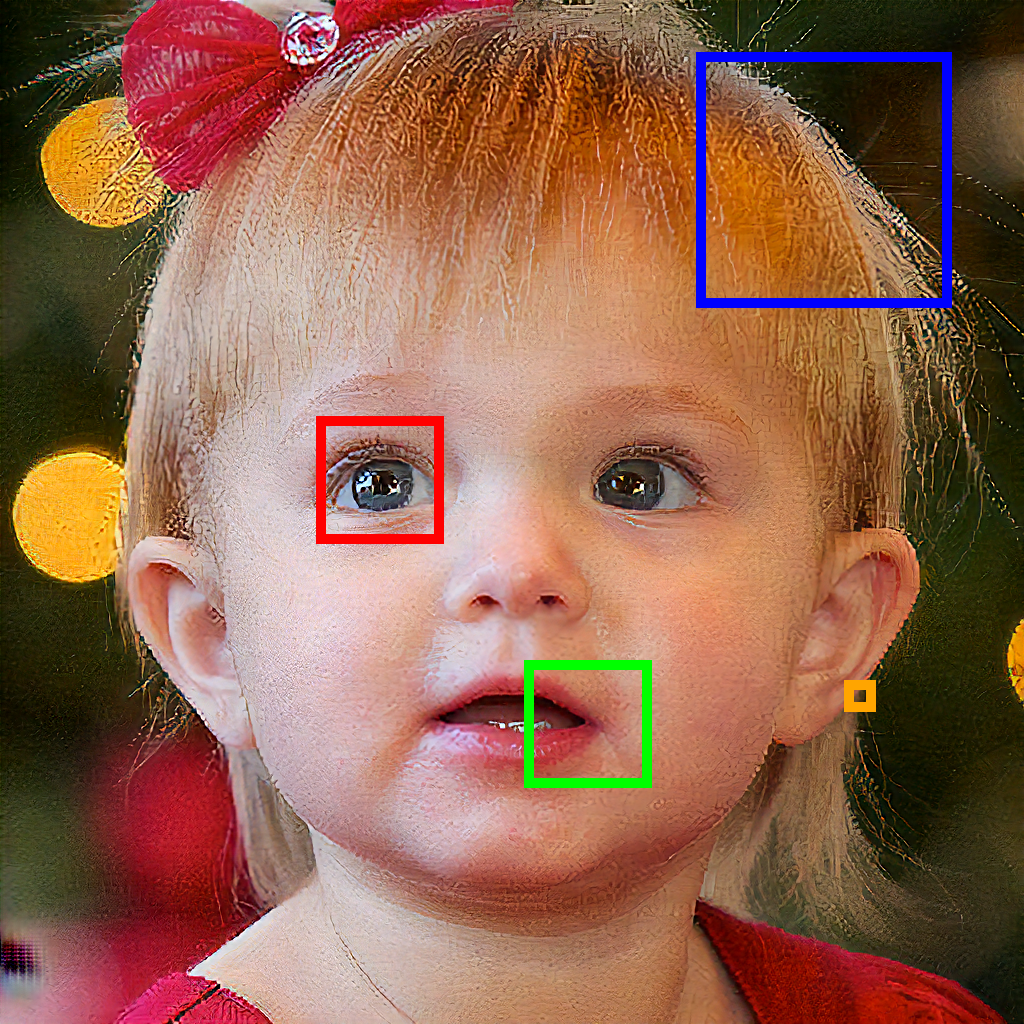}
        \\
        \includegraphics[width=\SizeFigCompareHRSmall\textwidth,cfbox=red 1pt 1pt]{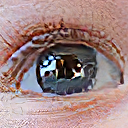}
        \includegraphics[width=\SizeFigCompareHRSmall\textwidth,cfbox=blue 1pt 1pt]{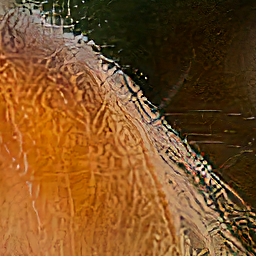}
        \\
        \includegraphics[width=\SizeFigCompareHRSmall\textwidth,cfbox=green 1pt 1pt]{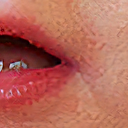}
        \includegraphics[width=\SizeFigCompareHRSmall\textwidth,cfbox=orange 1pt 1pt]{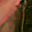}
        \caption{SRGAN~\cite{ledig2017photo} \\ (22.32 / 0.482)} 
        \label{fig:SRGAN}
    \end{subfigure}
    \\ 
    \begin{subfigure}[b]{\SizeFigCompareHRLarge\textwidth}
        \includegraphics[width=\textwidth]{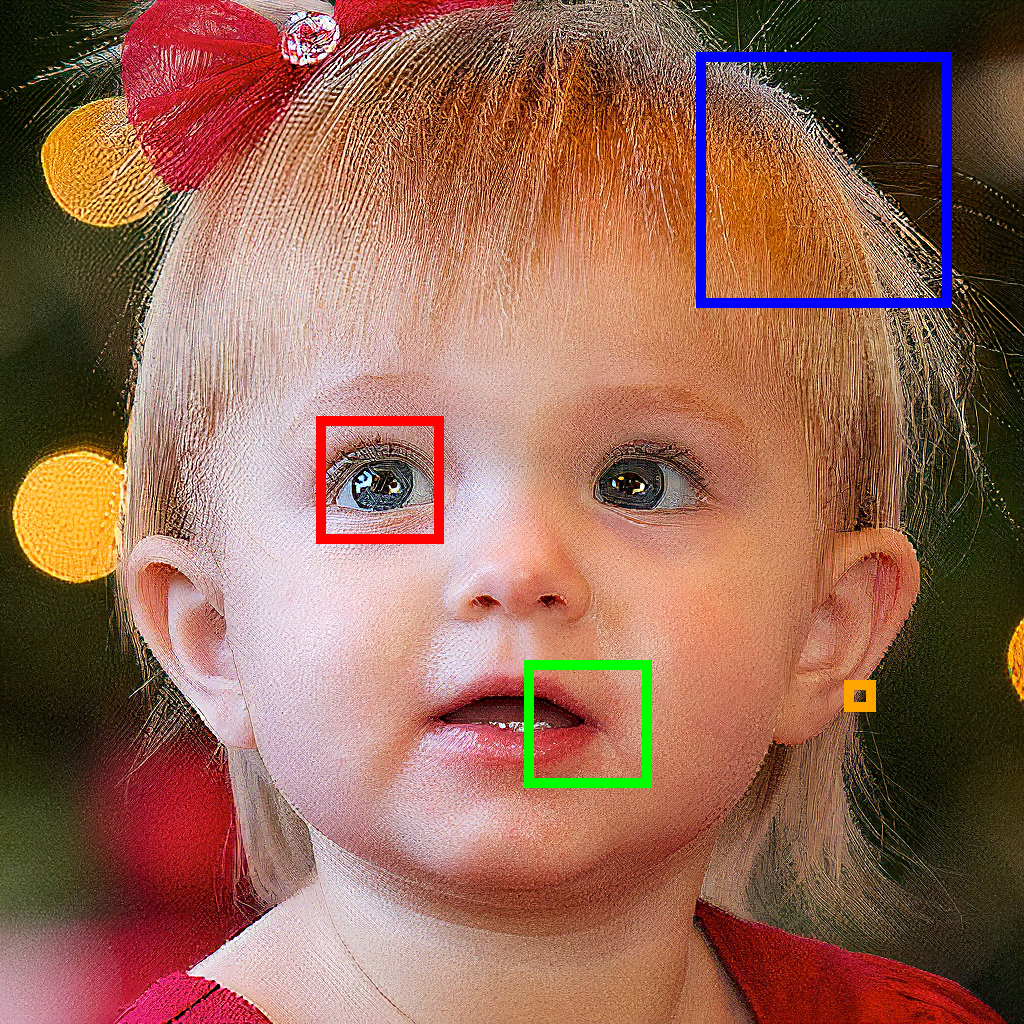}
        \\
        \includegraphics[width=\SizeFigCompareHRSmall\textwidth,cfbox=red 1pt 1pt]{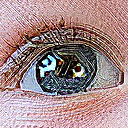}
        \includegraphics[width=\SizeFigCompareHRSmall\textwidth,cfbox=blue 1pt 1pt]{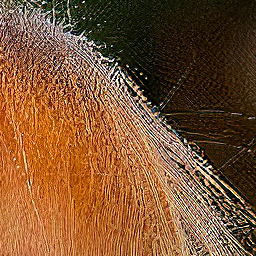}
        \\
        \includegraphics[width=\SizeFigCompareHRSmall\textwidth,cfbox=green 1pt 1pt]{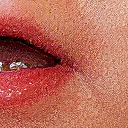}
        \includegraphics[width=\SizeFigCompareHRSmall\textwidth,cfbox=orange 1pt 1pt]{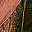}
        \caption{ESRGAN~\cite{wang2018esrgan} \\ (17.41 / 0.183)}
        \label{fig:ESRGAN}
    \end{subfigure}
    ~
    \begin{subfigure}[b]{\SizeFigCompareHRLarge\textwidth}
        \includegraphics[width=\textwidth]{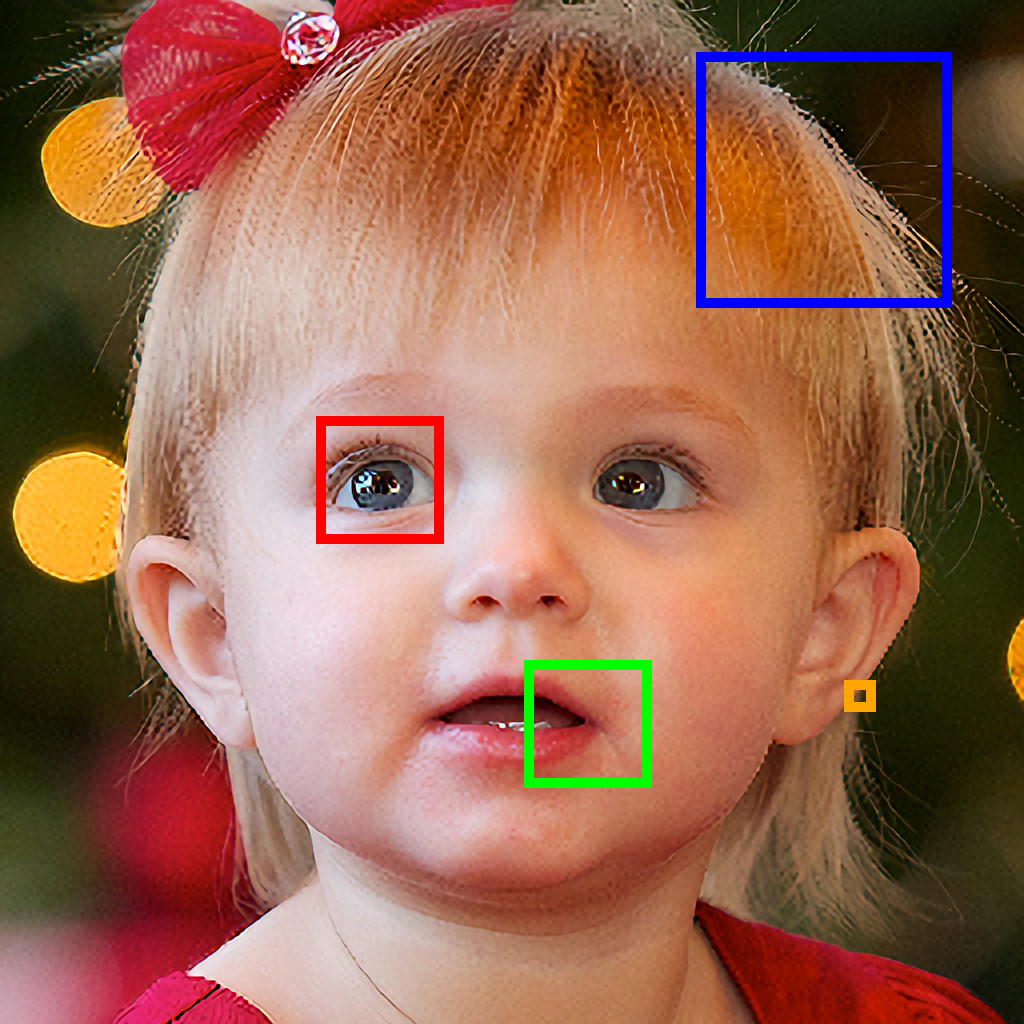}
        \\
        \includegraphics[width=\SizeFigCompareHRSmall\textwidth,cfbox=red 1pt 1pt]{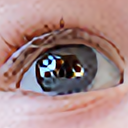}
        \includegraphics[width=\SizeFigCompareHRSmall\textwidth,cfbox=blue 1pt 1pt]{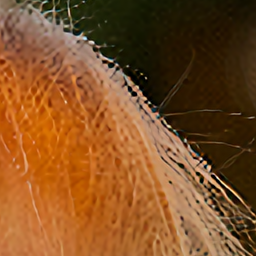}
        \\
        \includegraphics[width=\SizeFigCompareHRSmall\textwidth,cfbox=green 1pt 1pt]{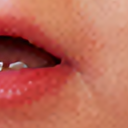}
        \includegraphics[width=\SizeFigCompareHRSmall\textwidth,cfbox=orange 1pt 1pt]{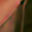}
        \caption{EDSR~\cite{lim2017enhanced} \\ (27.14 / 0.773)}
        \label{fig:EDSR}
    \end{subfigure}
    ~
    \begin{subfigure}[b]{\SizeFigCompareHRLarge\textwidth}
        \includegraphics[width=\textwidth]{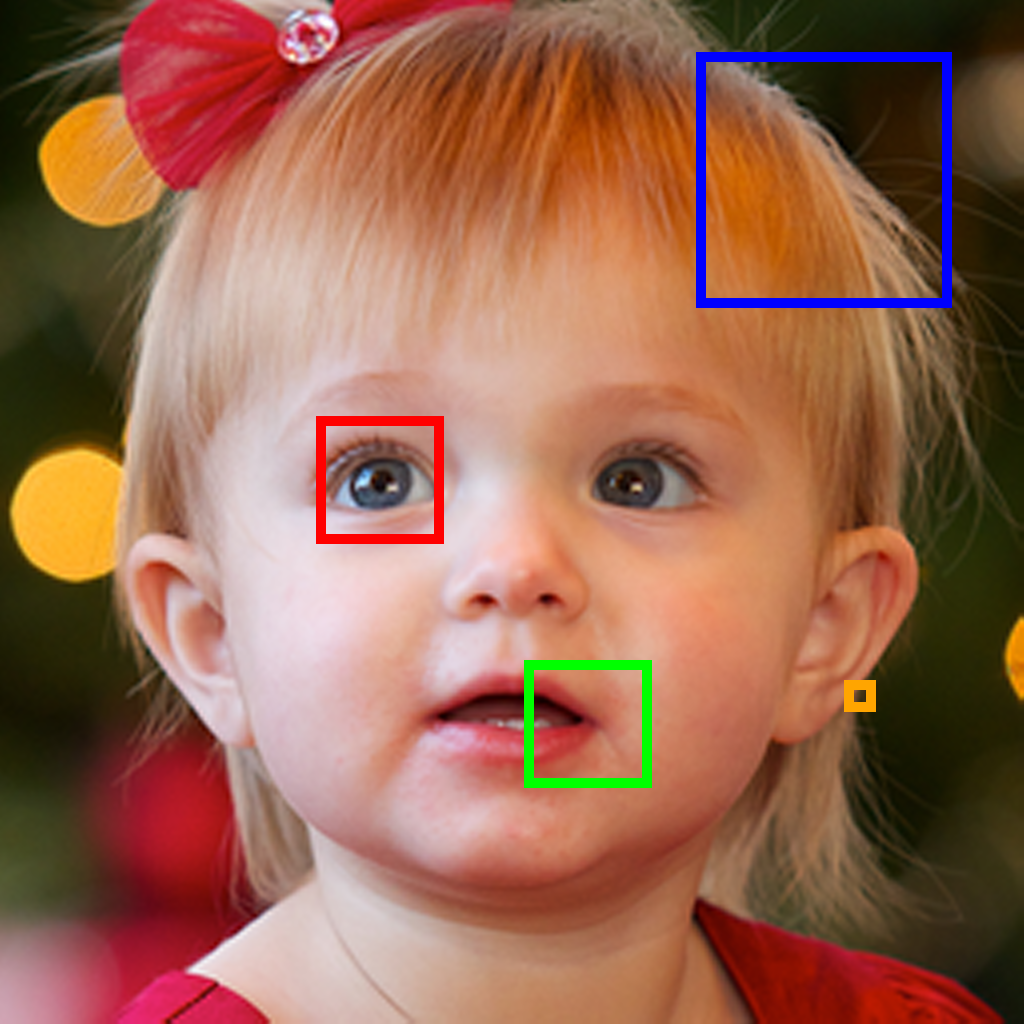}
        \\
        \includegraphics[width=\SizeFigCompareHRSmall\textwidth,cfbox=red 1pt 1pt]{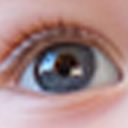}
        \includegraphics[width=\SizeFigCompareHRSmall\textwidth,cfbox=blue 1pt 1pt]{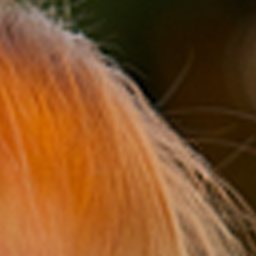}
        \\
        \includegraphics[width=\SizeFigCompareHRSmall\textwidth,cfbox=green 1pt 1pt]{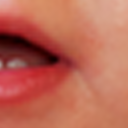}
        \includegraphics[width=\SizeFigCompareHRSmall\textwidth,cfbox=orange 1pt 1pt]{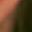}
        \caption{EnhanceNet~\cite{sajjadi2017enhancenet} \\ (29.24 / 0.799)}
        \label{fig:EnhanceNet}
    \end{subfigure}
    ~
    \begin{subfigure}[b]{\SizeFigCompareHRLarge\textwidth}
        \includegraphics[width=\textwidth]{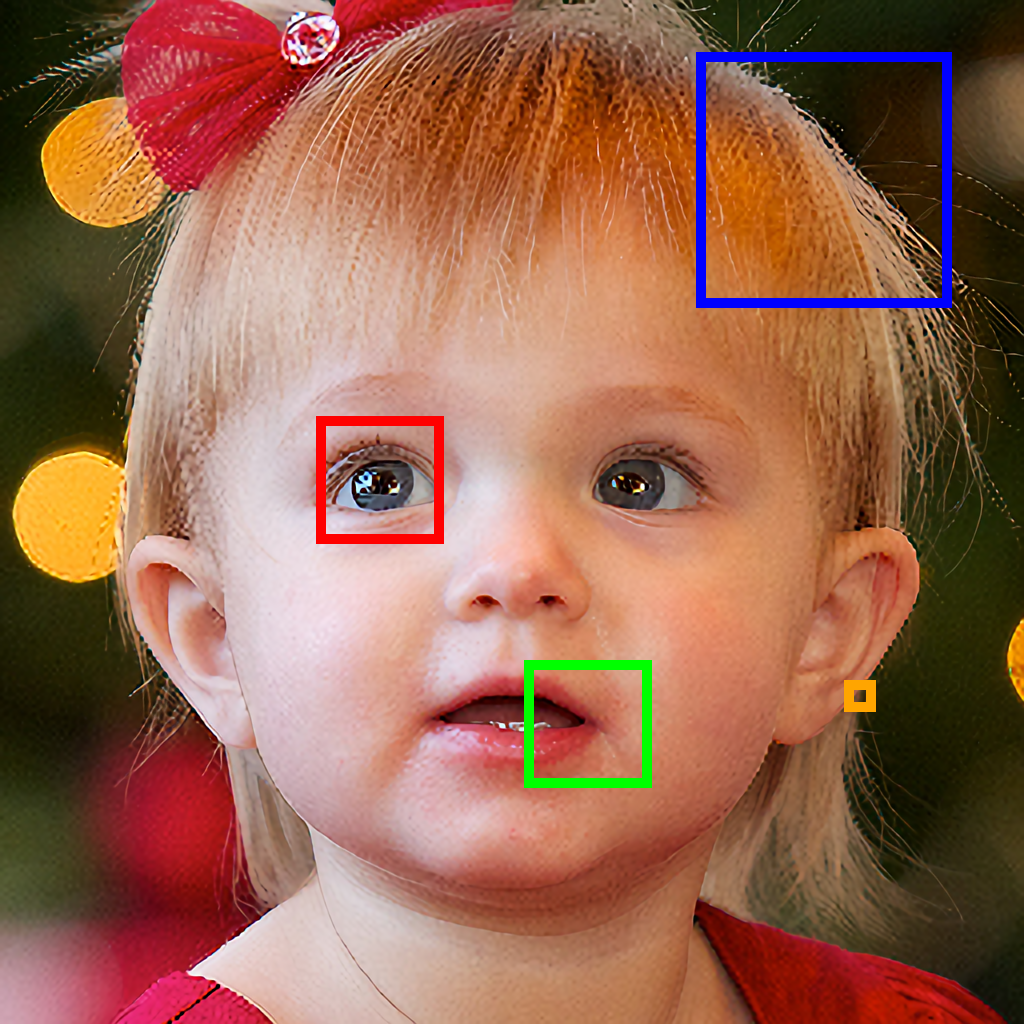}
        \\
        \includegraphics[width=\SizeFigCompareHRSmall\textwidth,cfbox=red 1pt 1pt]{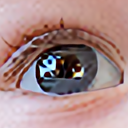}
        \includegraphics[width=\SizeFigCompareHRSmall\textwidth,cfbox=blue 1pt 1pt]{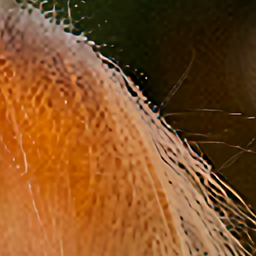}
        \\
        \includegraphics[width=\SizeFigCompareHRSmall\textwidth,cfbox=green 1pt 1pt]{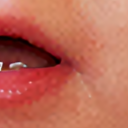}
        \includegraphics[width=\SizeFigCompareHRSmall\textwidth,cfbox=orange 1pt 1pt]{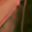}
        \caption{SRFBN~\cite{li2019feedback} \\ (26.65 / 0.765)}
        \label{fig:SRFBN}
    \end{subfigure}
    ~
    \begin{subfigure}[b]{\SizeFigCompareHRLarge\textwidth}
        \includegraphics[width=\textwidth]{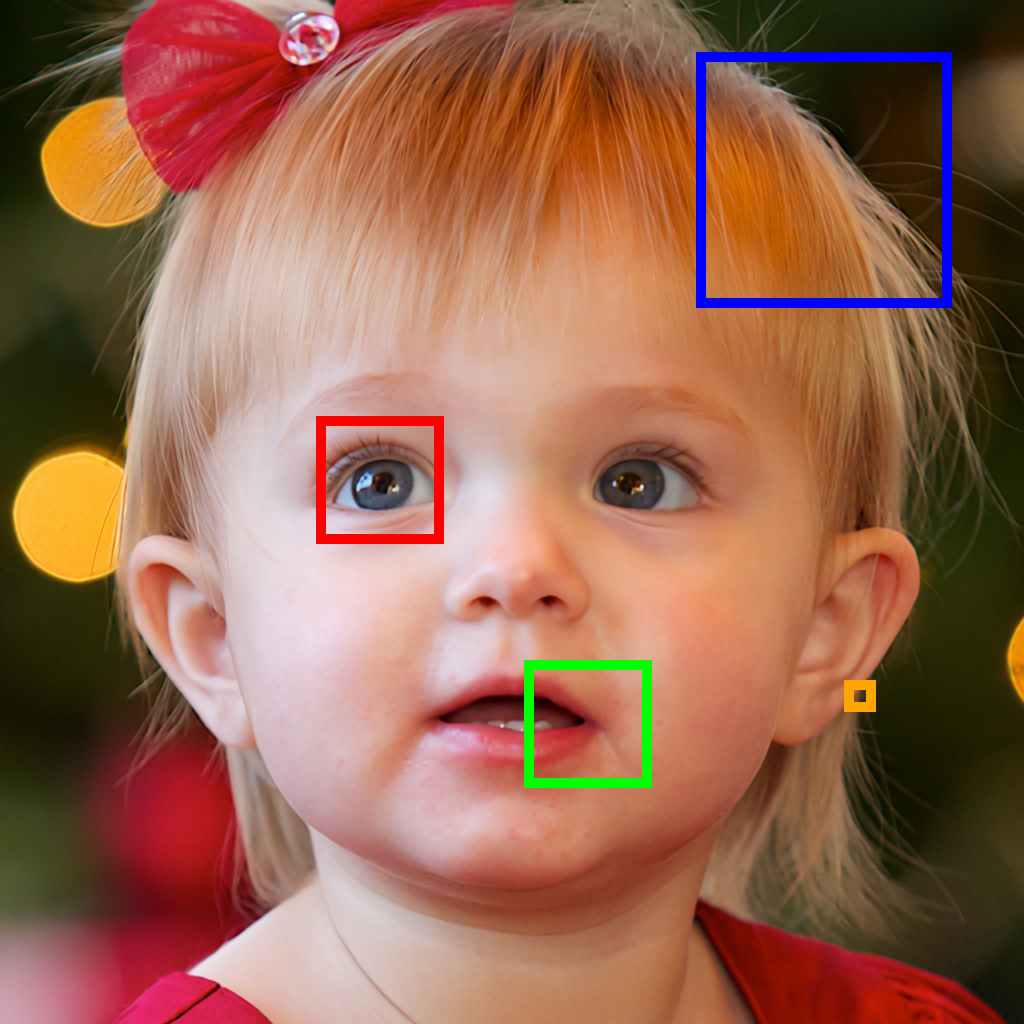}
        \\
        \includegraphics[width=\SizeFigCompareHRSmall\textwidth,cfbox=red 1pt 1pt]{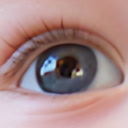}
        \includegraphics[width=\SizeFigCompareHRSmall\textwidth,cfbox=blue 1pt 1pt]{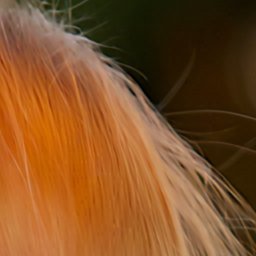}
        \\
        \includegraphics[width=\SizeFigCompareHRSmall\textwidth,cfbox=green 1pt 1pt]{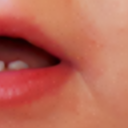}
        \includegraphics[width=\SizeFigCompareHRSmall\textwidth,cfbox=orange 1pt 1pt]{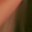}
        \caption{Ours \\ (33.92 / 0.893)}
        \label{fig:ours}
    \end{subfigure}
\end{center}
\vspace{-5mm}
\caption{Comparison with state-of-the-art methods for the \textbf{patch-based} version (output HR image is 1024$ \times$1024). As visible, our method can super-resolve without artifacts and noise-like patterns. Reconstructed images are visually pleasing and resemble the ground-truth better than the existing methods (for a better view, see in color on digital display). }
\label{fig:comparison}
\end{figure*}

\begin{figure*}[t]
\begin{center}
    \begin{subfigure}[b]{\SizeFigCompareLR\textwidth}
        \includegraphics[width=\textwidth]{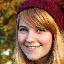}
        \includegraphics[width=\textwidth]{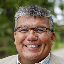}
        \includegraphics[width=\textwidth]{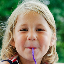}
        \includegraphics[width=\textwidth]{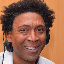}
        \caption{Input \\ (PSNR / SSIM)}
        \label{fig:input}
    \end{subfigure}
    \begin{subfigure}[b]{\SizeFigCompareLR\textwidth}
        \includegraphics[width=\textwidth]{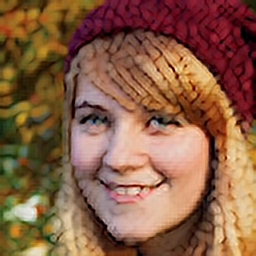}
        \includegraphics[width=\textwidth]{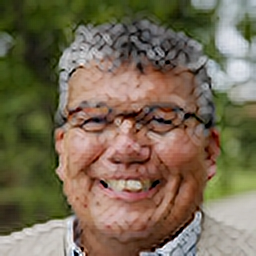}
        \includegraphics[width=\textwidth]{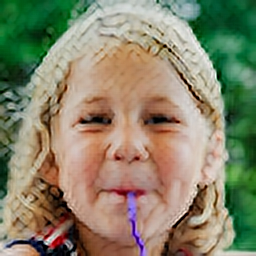}
        \includegraphics[width=\textwidth]{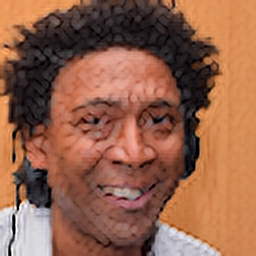}
        \caption{SRCNN~\cite{dong2015image} \\ (22.82 / 0.668)}
        \label{fig:SRCNN_256}
    \end{subfigure}
    \begin{subfigure}[b]{\SizeFigCompareLR\textwidth}
        \includegraphics[width=\textwidth]{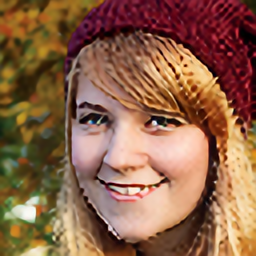}
        \includegraphics[width=\textwidth]{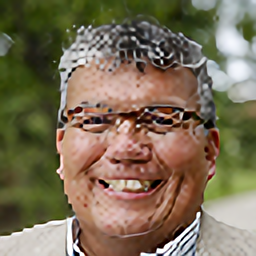}
        \includegraphics[width=\textwidth]{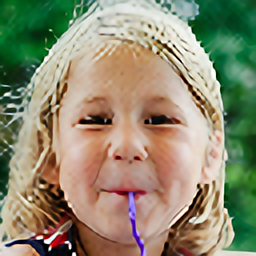}
        \includegraphics[width=\textwidth]{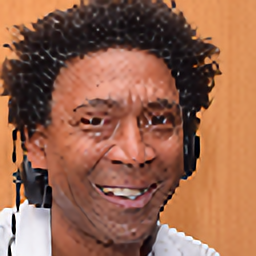}
        \caption{EDSR~\cite{lim2017enhanced} \\ (21.78 / 0.689)}
        \label{fig:EDSR_256}
    \end{subfigure}
    \begin{subfigure}[b]{\SizeFigCompareLR\textwidth}
        \includegraphics[width=\textwidth]{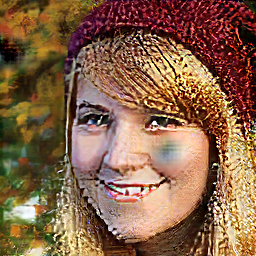}
        \includegraphics[width=\textwidth]{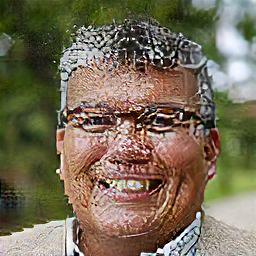}
        \includegraphics[width=\textwidth]{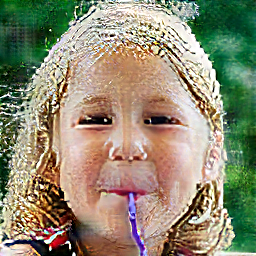}
        \includegraphics[width=\textwidth]{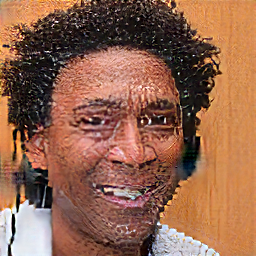}
        \caption{SRGAN~\cite{ledig2017photo} \\ (17.48 / 0.420)}
        \label{fig:SRGAN_256}
    \end{subfigure}
    \begin{subfigure}[b]{\SizeFigCompareLR\textwidth}
        \includegraphics[width=\textwidth]{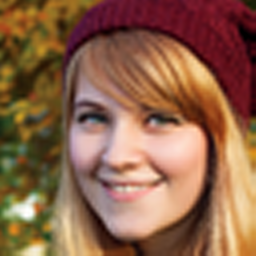}
        \includegraphics[width=\textwidth]{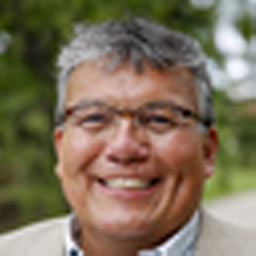}
        \includegraphics[width=\textwidth]{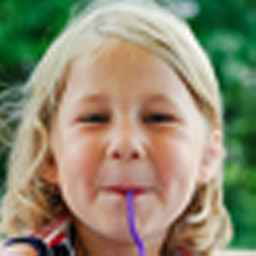}
        \includegraphics[width=\textwidth]{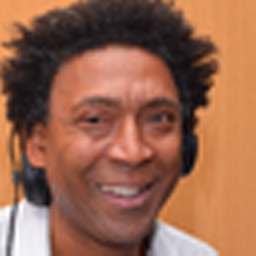}
        \caption{E-Net~\cite{sajjadi2017enhancenet} \\ (23.08 / 0.679)}
        \label{fig:EnhanceNet_256}
    \end{subfigure}
    \begin{subfigure}[b]{\SizeFigCompareLR\textwidth}
        \includegraphics[width=\textwidth]{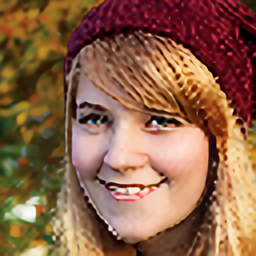}
        \includegraphics[width=\textwidth]{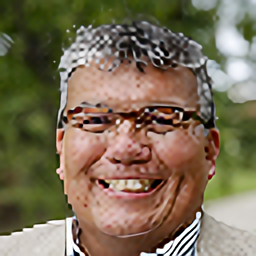}
        \includegraphics[width=\textwidth]{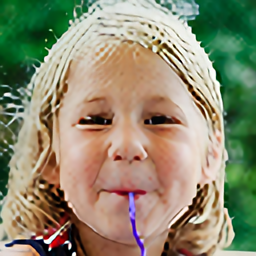}
        \includegraphics[width=\textwidth]{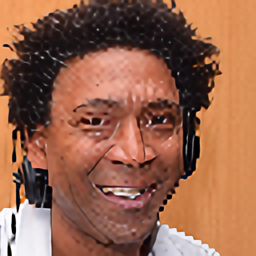}
        \caption{SRFBN~\cite{li2019feedback} \\ (21.12 / 0.673)}
        \label{fig:SRFBN_256}
    \end{subfigure}
    \begin{subfigure}[b]{\SizeFigCompareLR\textwidth}
        \includegraphics[width=\textwidth]{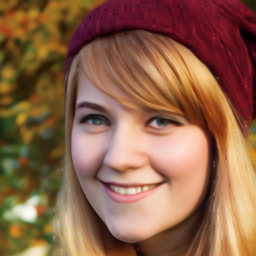}
        \includegraphics[width=\textwidth]{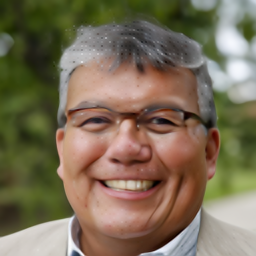}
        \includegraphics[width=\textwidth]{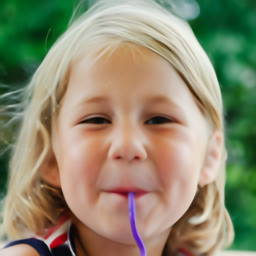}
        \includegraphics[width=\textwidth]{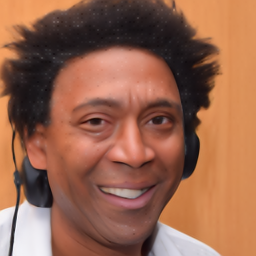}
        \caption{Ours \\ (26.79 / 0.800)}
        \label{fig:ours_256}
    \end{subfigure}
\end{center}
\vspace{-5mm}
\caption{Comparisons with the state-of-the-art for the \textbf{whole-face} version, i.e. training with 64$\times$64 face images as input to generate 4$\times$ HR outputs of size 256$\times$256. As visible, our method generates superior results for whole-face training as well. 
}
\label{fig:comparison_256}
\end{figure*}

\section{Experiments}\label{sec:experiment}

\subsection{Dataset} 

We use 1024$\times$1024 face images from the Flickr-Faces-HQ Dataset (FFHQ)~\cite{karras2019style}, which consists of 70,000 high-quality PNG images with considerable variation in terms of facial attributes such as age and ethnicity as well as image background. It also provides sufficient coverage of accessories such as eyeglasses, sunglasses, and hats. The images were crawled from Flickr. We then randomly split the FFHQ dataset into non-overlapping training, testing, and validation subsets of ratio 80\%, 15\%, and 5\%, respectively.

\subsection{Evaluation Metrics}\label{sec:metric}

To quantitatively measure the performance and provide comprehensive comparisons with state-of-the-art methods, we used four quality assessment metrics including PSNR, SSIM, FID, and MS-SSIM. FID~\cite{heusel2017gans} is defined as:
\begin{equation}
  \text{FID} = ||\mu_r - \mu_g||^2 + \text{Tr} (\Sigma_r + \Sigma_g - 2 (\Sigma_r \Sigma_g)^{1/2}),
\end{equation}
where $X_r \sim \mathcal{N} (\mu_r, \Sigma_r)$ and $X_g \sim \mathcal{N} (\mu_g, \Sigma_g)$ are activations of Inception-v3 pool3 layer for real and generated samples, respectively. Lower FIDs mean the generated results are closer to the original image, measured by the Fr\'echet distance between two distributions.

\subsection{Comparisons}

We quantitatively compare our method with seven state-of-the-art super-resolution approaches as well as with the bicubic upsampling. 

In the inference (test) time, we can process the given image either by taking it as a whole (whole-face) or by patch-by-patch (patch-based). The memory limitations of GPUs set an upper bound on the input image size, in particular for the training phase.  For example, memory limitations of the single GPU we used prohibited training with 1024$\times$1024 input images. Thus, to train with and infer from such relatively large images (e.g., 1024$\times$1024), we employ the patch-based version. For these two alternative versions, we trained separate models:
\begin{itemize}
\item The whole-face version uses the entire face image as an input. One can argue that using the face as a whole would provide better semantics. In our experiments, we set the input size to 256$\times$256 face images. 
\item The patch-based version uses the same network as above and identical size overlapping patches. We set the patch size to 256$\times$256 for 1024$\times$1024 HR outputs. We also tested 128$\times$128 patch size. The patch-based version allows generating very large output faces without being restricted by the GPU memory.  
\end{itemize}
The only difference between the above versions is the training data. For training of the patch-based version, we sampled randomly sampled around 2 million patches (48 per image) from 56,000 HR training images. For both versions, we used conventional 4$\times$ bicubic downsampling to obtain the LR input image. We augmented the training data by applying one of these six geometric transformations; rotating 90$^{\circ}$, 180$^{\circ}$, 270$^{\circ}$, flipping vertically, and flipping horizontally. Even for the whole-face version, the local receptive fields do not derive semantics from the entire face. Supporting the assumption that a network would be as best as its local constituent kernels, we observed that the patch-based version generates competitive models while accessing finer granularity textures. 

Figures~\ref{fig:comparison} and \ref{fig:comparison_256} provide qualitative comparisons with state-of-the-art super-resolution techniques for 1024$\times$1024 images (patch-based) and 256$\times$256 images (whole-face), respectively. We used the best models available for the state-of-the-art methods provided by their authors. Our results showcase the superior quality of the proposed method. In particular, our patch-based version achieves the most pleasing HR reconstructions without any artifacts. In comparison, the GAN-based methods introduce perceptually unignorable fragmentations, remnant noise-like patterns, and broken textures. We also provide quantitative results in Tables~\ref{tab:comparison} and \ref{tab:comparison_256}, where our model outperforms the compared state-of-the-art methods with a remarkable margin under various metrics, including PSNR, SSIM, and FID. Notice that the compared methods either do not take advantage of facial semantic information or impose incorrect semantic bias. They do not use facial components to guide the super-resolution process. Most use lower resolution images in their training, which may be further limiting their representation capacity. These explain why bicubic upsampling, a deterministic approach without any semantic bias, performs better than its data-driven counterparts.

\begin{table}[]
\centering
\resizebox{0.475\textwidth}{!}{
\begin{tabular}{@{}lcccc@{}}
\toprule
                                & PSNR & SSIM & MS-SSIM & FID \\ \midrule
Bicubic                         &  31.87             &  0.872             &  0.956                              &  \textbf{10.65}            \\
SRCNN~\cite{dong2015image}      &  27.40             &  0.801             &  0.924                             &  31.84            \\
FSRCNN~\cite{dong2016accelerating}& 24.71             &  0.804             &  0.951                                      &  23.97            \\
EDSR~\cite{lim2017enhanced}     &  28.34             &  0.827             &  0.933                                 &  15.54            \\
SRGAN~\cite{ledig2017photo}     &  21.49             &  0.515             &  0.807                                    &  60.67            \\
ESRGAN~\cite{wang2018esrgan}    &  19.84             &  0.353             &  0.782                                     &  72.73            \\
EnhanceNet~\cite{sajjadi2017enhancenet}&  29.42      &  0.832             &  0.934                                    &  19.07            \\
SRFBN~\cite{li2019feedback}     &  27.90             &  0.822             &  0.931                                      &  17.14             \\
Ours                 &  \textbf{34.10}                        &  \textbf{0.906}             &  \textbf{0.971}                                       &  12.40            \\
\bottomrule
\end{tabular}
}
\caption{Comparison results for 1024$\times$1024 outputs. Our method is trained with \textbf{patches}.}
\label{tab:comparison}
\end{table}

\begin{table}[]
\centering
\resizebox{0.475\textwidth}{!}{
\begin{tabular}{@{}lcccc@{}}
\toprule
                                & PSNR & SSIM & MS-SSIM & FID \\ \midrule
Bicubic                         &  25.57             &  0.766             &  0.935                                      &  135.51           \\
SRCNN~\cite{dong2015image}      &  23.12             &  0.688             &  0.900                                     &  147.21           \\
FSRCNN~\cite{dong2016accelerating}& 22.45             &  0.709             &  0.930                                      &  139.78           \\
EDSR~\cite{lim2017enhanced}     &  22.47             &  0.706             &  0.901                                      &  129.14           \\
SRGAN~\cite{ledig2017photo}     &  17.57             &  0.415             &  0.757                                       &  156.07           \\
ESRGAN~\cite{wang2018esrgan}    &  15.43             &  0.267             &  0.747                                       &  166.36           \\
EnhanceNet~\cite{sajjadi2017enhancenet}&  23.64      &  0.701             &  0.897                                       &  116.38           \\
SRFBN~\cite{li2019feedback}     &  21.96             &  0.693             &  0.895                                      &  132.59           \\
Ours                            &  \textbf{27.42}             &  \textbf{0.816}             &  \textbf{0.958}                                      &  \textbf{74.43}            \\
\bottomrule
\end{tabular}
}
\caption{Comparison results for 256$\times$256 outputs. Our method is trained with \textbf{whole-faces}.}
\label{tab:comparison_256}
\end{table}

\begin{figure}
\centering
  \includegraphics[width=0.95\linewidth]{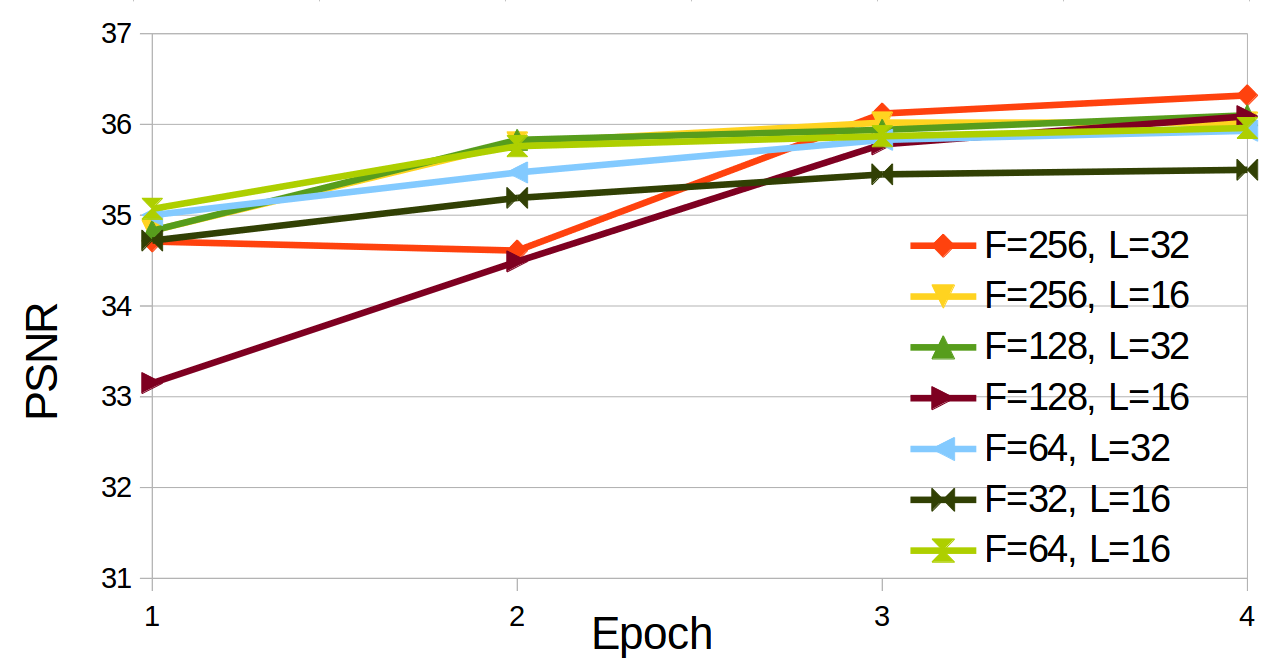}
  \vspace{-2mm}
  \caption{Effect of different network parameters on accuracy on the test dataset at initial epochs during the training phase. F is the number of features in each layer and L is the number of layers. PSNR is in dB. As visible, most versions converge to the higher PSNR scores quickly in a few epochs. This shows that our network is robust to the different hyper-parameterizations.}
  \label{fig:ab1}
\end{figure}

\begin{figure}
\centering
  \includegraphics[width=0.6\linewidth]{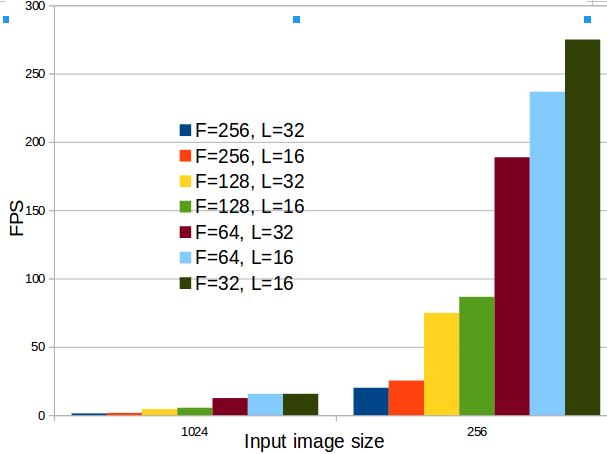}
    \vspace{-2mm}
  \caption{Effect of different network parameters on speed.}
  \label{fig:ab2}
\end{figure}

\subsection{Ablation Study}

We evaluated variants of our patch-wise model with different hyper-parameterizations, \ie, $F$, the number of features per layer, and $L$, the number of Resblock layers. Figure~\ref{fig:ab1} demonstrates the training performance in terms of the attained PSNR scores after the initial training epochs of different configurations. As expected, with the increasing number of features and Resblock layers the performance gets better. 
Most of the progress towards convergence were achieved in the initial epochs when we trained the final model using an NVIDIA DGX-1.
As shown in Figure~\ref{fig:ab1}, the network trained with different hyperparameters converges to a similar level after a few epochs, indicating that the proposed network is generally applicable regardless of the settings of $F$ and $L$, thus can adapt in accordance with real world scenarios such as GPU memory limit.

We also analyzed the performance of using a single stage 4$\times$ super-resolution instead of two-stage network. This version, even though attained better scores than the compared state-of-the-art methods, could not reach our two-stage PSNR performance: 33.71 dB (single stage) vs. 34.10 dB (two stage) for the patch-based version, and 26.46 dB (single stage) vs. 27.42 dB (two stage) for the whole-face version. 

Figure~\ref{fig:ab2} compares the inference speed of our model achieved for both 1024$\times$1024 (patch-based) and 256$\times$256 (whole-face) image resolutions on a GTX 2080Ti GPU at the 4$\times$ super-resolution setting. We can attain 270 fps on the whole 256$\times$256 image size and 15 fps of 1024$\times$1024 in typical settings. The whole-face processing sets an upper bound on the overall latency of the model for the parallel-processing platforms while significantly exceeding real-time speed. Our patch-based solution can reach 15 fps and substantially improves the PSNR scores. It is possible to attain real-time performance for the patch-based version by sampling the input patches from the input image with lesser degrees of overlaps.

\section{Conclusion}\label{sec:conclusion}
We show that imposing attention maps implicitly and regularizing the super-resolution process using loss functions from intermediate and final upscaling stages significantly improves the performance as demonstrated in our superior results. Our patch-based method has the advantage of processing any input size image. As future work, we plan to train the entire network, including the component segmentation part in an end-to-end fashion. 

{\small
\bibliographystyle{ieee}
\bibliography{db}
}

\newpage
\clearpage
\appendix
\section{Supplementary Materials}

\begin{figure*}[b!]
\begin{center}
\includegraphics[width=0.225\textwidth]{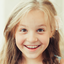}~\includegraphics[width=0.225\textwidth]{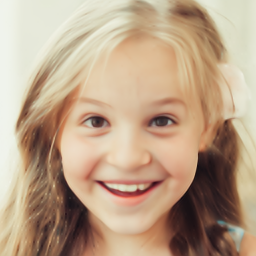}
~
\includegraphics[width=0.225\textwidth]{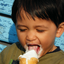}~\includegraphics[width=0.225\textwidth]{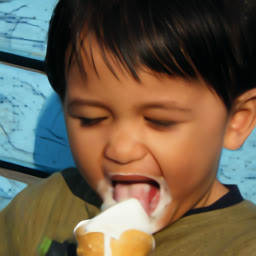}
~
\includegraphics[width=0.225\textwidth]{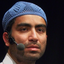}~\includegraphics[width=0.225\textwidth]{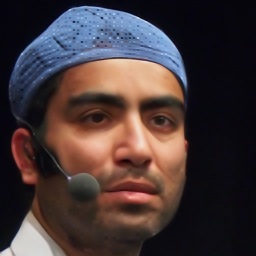}
~
\includegraphics[width=0.225\textwidth]{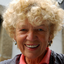}~\includegraphics[width=0.225\textwidth]{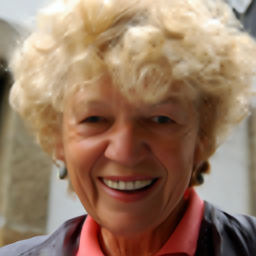}
~
\includegraphics[width=0.225\textwidth]{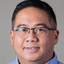}~\includegraphics[width=0.225\textwidth]{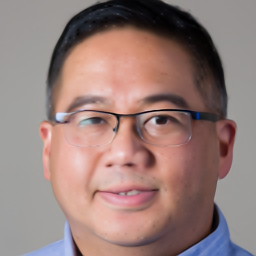}
~
\includegraphics[width=0.225\textwidth]{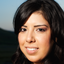}~\includegraphics[width=0.225\textwidth]{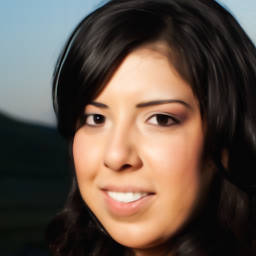}
~
\includegraphics[width=0.225\textwidth]{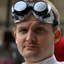}~\includegraphics[width=0.225\textwidth]{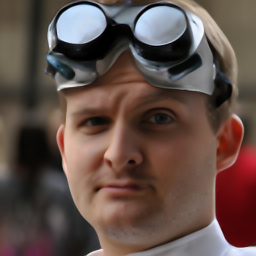}
~
\includegraphics[width=0.225\textwidth]{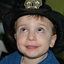}~\includegraphics[width=0.225\textwidth]{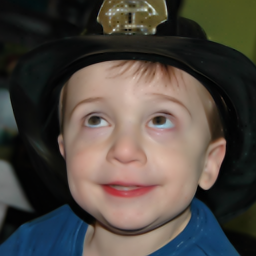}
~
\includegraphics[width=0.225\textwidth]{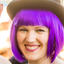}~\includegraphics[width=0.225\textwidth]{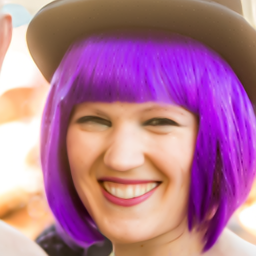}
~
\includegraphics[width=0.225\textwidth]{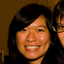}~\includegraphics[width=0.225\textwidth]{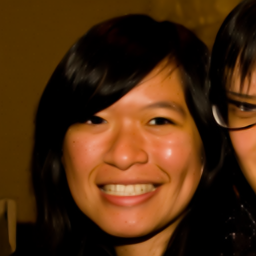}
\end{center}
\caption{4$\times$ super-resolution with our \textbf{whole-face} method. In each pair, input image is 64$\times$64 and output image is 256$\times$256. Please view on digital display for the best view.}
\label{fig:comparison_256_appendix}
\end{figure*}

\begin{figure*}[th]
\begin{center}
    \begin{subfigure}[b]{0.4 \textwidth}
        \includegraphics[width=\textwidth]{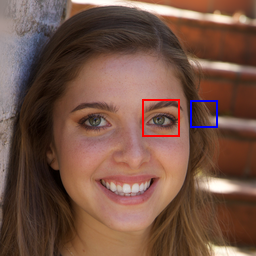}
        \\
        \includegraphics[width=0.980\textwidth,cfbox=red 1pt 1pt]{img/00736_input_RED.png}
        \includegraphics[width=0.980\textwidth,cfbox=blue 1pt 1pt]{img/00736_input_BLUE.png}
        \caption{Input image  and zoomed in regions.}
    \end{subfigure}
    \begin{subfigure}[b]{0.4 \textwidth}
        \includegraphics[width=\textwidth]{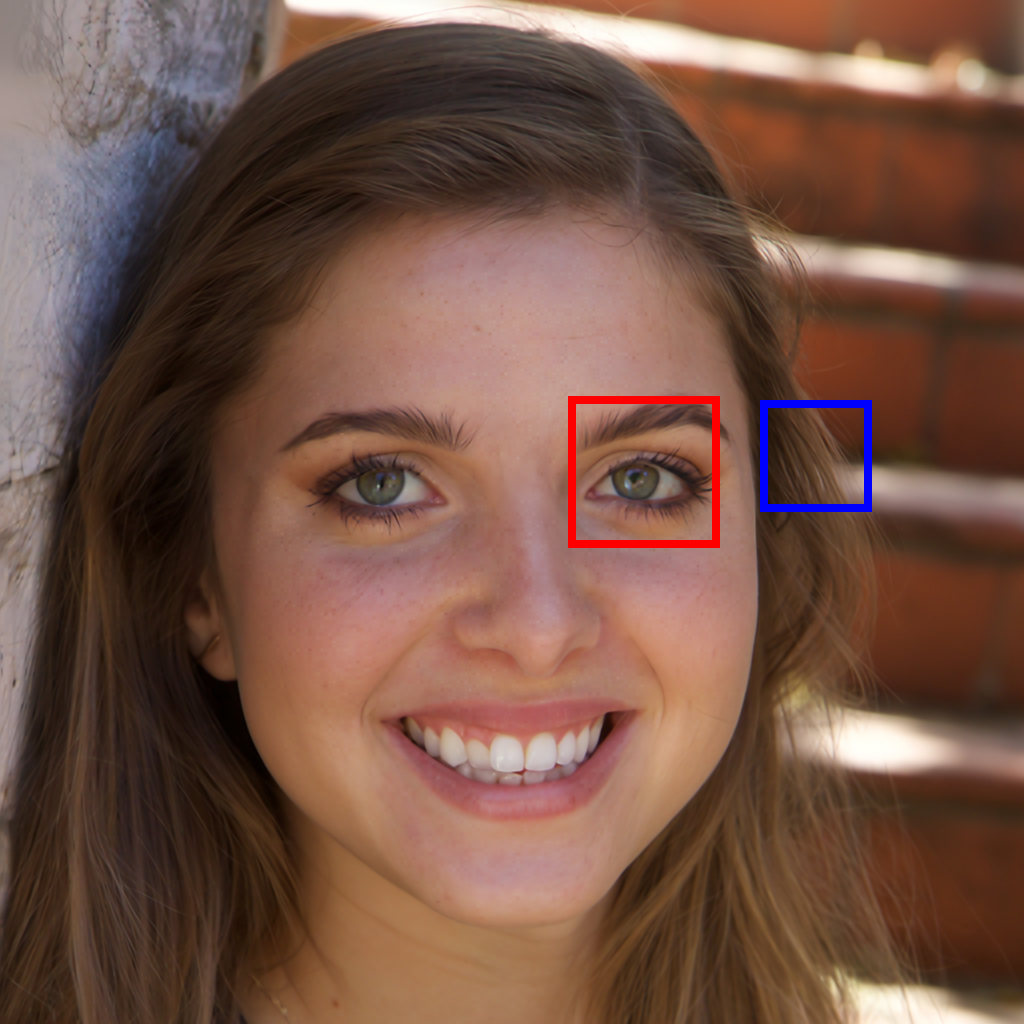}
        \\
        \includegraphics[width=0.980\textwidth,cfbox=red 1pt 1pt]{img/00736_ours_plus_mask_RED.png}
        \includegraphics[width=0.980\textwidth,cfbox=blue 1pt 1pt]{img/00736_ours_plus_mask_BLUE.png}
        \caption{Our result.}
    \end{subfigure}
\end{center}
\caption{4$\times$ super-resolution with our \textbf{patch-based} method. Input image is 256$\times$256, output image is 1024$\times$1024. Please view on digital display for the best view.}
\label{fig:appendix}
\end{figure*}

\begin{figure*}[th]
\begin{center}
    \begin{subfigure}[b]{0.4 \textwidth}
        \includegraphics[width=\textwidth]{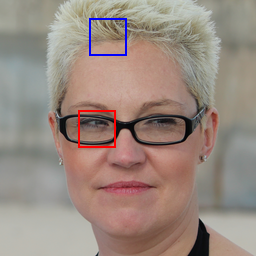}
        \\
        \includegraphics[width=0.980\textwidth,cfbox=red 1pt 1pt]{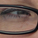}
        \includegraphics[width=0.980\textwidth,cfbox=blue 1pt 1pt]{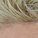}
        \caption{Input image  and zoomed in regions.}
    \end{subfigure}
    \begin{subfigure}[b]{0.4 \textwidth}
        \includegraphics[width=\textwidth]{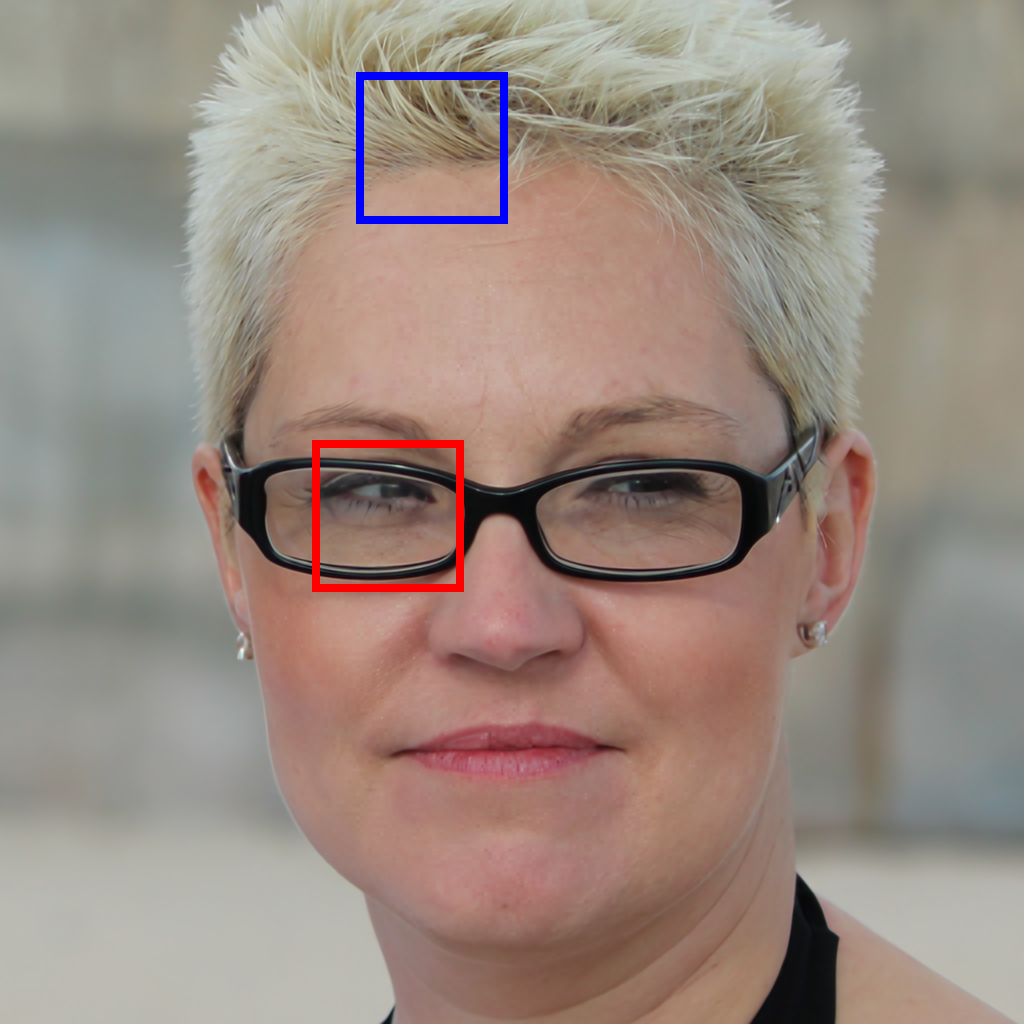}
        \\
        \includegraphics[width=0.980\textwidth,cfbox=red 1pt 1pt]{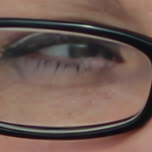}
        \includegraphics[width=0.980\textwidth,cfbox=blue 1pt 1pt]{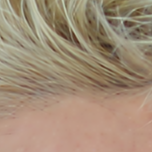}
        \caption{Our result.}
    \end{subfigure}
\end{center}
\caption{4$\times$ super-resolution with our \textbf{patch-based} method. Input image is 256$\times$256, output image is 1024$\times$1024. Please view on digital display for the best view.}
\label{fig:appendix}
\end{figure*}

\begin{figure*}[th]
\begin{center}
    \begin{subfigure}[b]{0.4 \textwidth}
        \includegraphics[width=\textwidth]{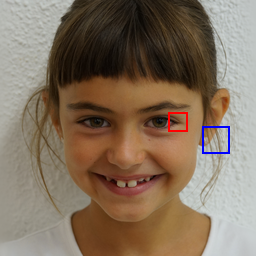}
        \\
        \includegraphics[width=0.980\textwidth,cfbox=red 1pt 1pt]{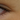}
        \includegraphics[width=0.980\textwidth,cfbox=blue 1pt 1pt]{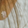}
        \caption{Input image  and zoomed in regions.}
    \end{subfigure}
    \begin{subfigure}[b]{0.4 \textwidth}
        \includegraphics[width=\textwidth]{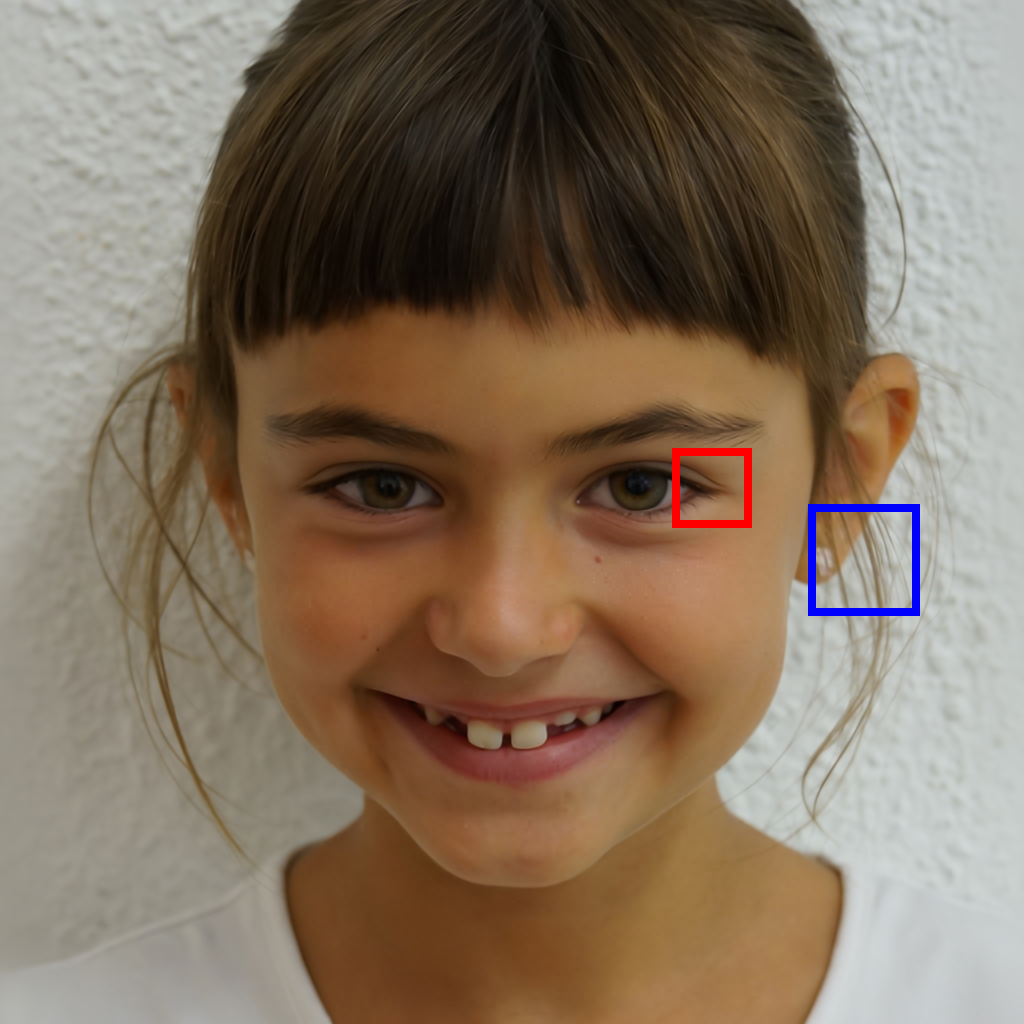}
        \\
        \includegraphics[width=0.980\textwidth,cfbox=red 1pt 1pt]{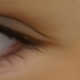}
        \includegraphics[width=0.980\textwidth,cfbox=blue 1pt 1pt]{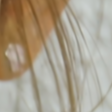}
        \caption{Our result.}
    \end{subfigure}
\end{center}
\caption{4$\times$ super-resolution with our \textbf{patch-based} method. Input image is 256$\times$256, output image is 1024$\times$1024. Please view on digital display for the best view.}
\label{fig:appendix}
\end{figure*}

\begin{figure*}[th]
\begin{center}
    \begin{subfigure}[b]{0.4 \textwidth}
        \includegraphics[width=\textwidth]{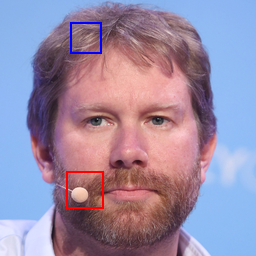}
        \\
        \includegraphics[width=0.980\textwidth,cfbox=red 1pt 1pt]{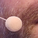}
        \includegraphics[width=0.980\textwidth,cfbox=blue 1pt 1pt]{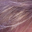}
        \caption{Input image  and zoomed in regions.}
    \end{subfigure}
    \begin{subfigure}[b]{0.4 \textwidth}
        \includegraphics[width=\textwidth]{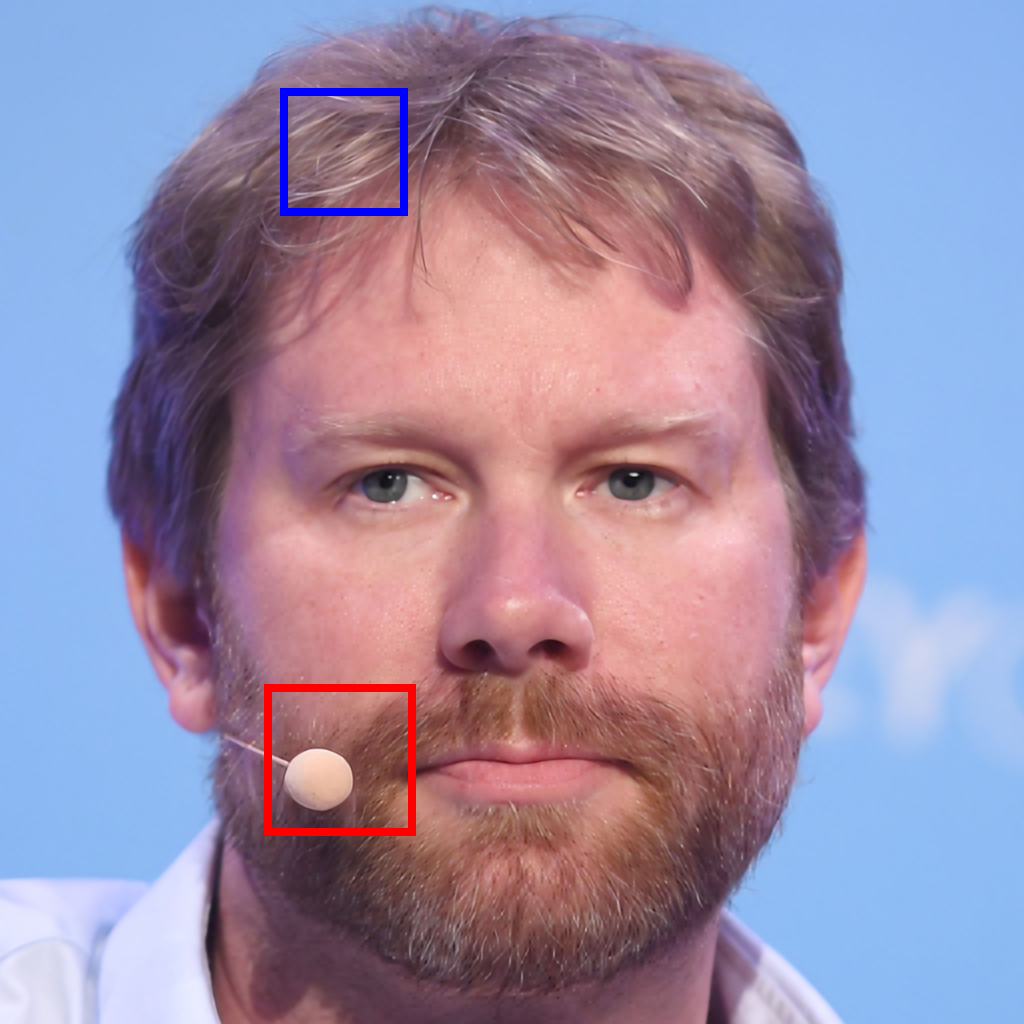}
        \\
        \includegraphics[width=0.980\textwidth,cfbox=red 1pt 1pt]{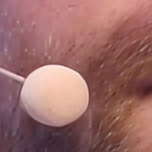}
        \includegraphics[width=0.980\textwidth,cfbox=blue 1pt 1pt]{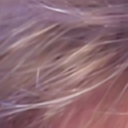}
        \caption{Our result.}
    \end{subfigure}
\end{center}
\caption{4$\times$ super-resolution with our \textbf{patch-based} method. Input image is 256$\times$256, output image is 1024$\times$1024. Please view on digital display for the best view.}
\label{fig:appendix}
\end{figure*}

\end{document}